\documentclass[10pt,twocolumn,letterpaper]{article}

\usepackage{color,xcolor}
\usepackage{epsfig}
\usepackage{graphicx}
\usepackage{subfiles}

\usepackage{adjustbox}
\usepackage{array}
\usepackage{booktabs}
\usepackage{colortbl}
\usepackage{float,wrapfig}
\usepackage{hhline}
\usepackage{multirow}
\usepackage{subcaption} %
\usepackage[labelfont={bf},labelsep={period},font={small}]{caption}

\let\llncssubparagraph\subparagraph
\let\subparagraph\paragraph
\usepackage[compact]{titlesec}
\let\subparagraph\llncssubparagraph

\usepackage{amsmath,amsfonts,amssymb}
\usepackage{bm}
\usepackage{nicefrac}
\usepackage{microtype}

\usepackage{changepage}
\usepackage{extramarks}
\usepackage{fancyhdr}
\usepackage{lastpage}
\usepackage{setspace}
\usepackage{soul}
\usepackage{xspace}

\usepackage{url}

\usepackage{algorithm, algorithmic}
\usepackage{enumerate}

\usepackage{iccv}
\usepackage{times}

\usepackage[pagebackref=true,breaklinks=true,letterpaper=true,colorlinks,bookmarks=false]{hyperref}

\usepackage{pbox}
\usepackage{footnote}
\usepackage{tablefootnote}

\usepackage{paralist}

\usepackage{enumitem}
\setitemize{noitemsep,topsep=0pt,parsep=0pt,partopsep=0pt}
\newcolumntype{L}[1]{>{\raggedright\let\newline\\\arraybackslash\hspace{0pt}}m{#1}}
\newcolumntype{C}[1]{>{\centering\let\newline\\\arraybackslash\hspace{0pt}}m{#1}}
\newcolumntype{R}[1]{>{\raggedleft\let\newline\\\arraybackslash\hspace{0pt}}m{#1}}

\newcommand{\ignorethis}[1]{}

\makeatletter
\DeclareRobustCommand\onedot{\futurelet\@let@token\@onedot}
\def\@onedot{\ifx\@let@token.\else.\null\fi\xspace}

\def\ie{\emph{i.e}\onedot} 
 
\def\etc{\emph{etc}\onedot} 
 
\def\etal{\emph{et al}\onedot}
\makeatother

\definecolor{citecolor}{RGB}{34,139,34}
\definecolor{mydarkblue}{rgb}{0,0.08,1}
\definecolor{mydarkgreen}{rgb}{0.02,0.6,0.02}
\definecolor{mydarkred}{rgb}{0.8,0.02,0.02}
\definecolor{mydarkorange}{rgb}{0.40,0.2,0.02}
\definecolor{mypurple}{RGB}{111,0,255}
\definecolor{myred}{rgb}{1.0,0.0,0.0}
\definecolor{mygold}{rgb}{0.75,0.6,0.12}
\definecolor{mydarkgray}{rgb}{0.66, 0.66, 0.66}

\def\multirowcenter{-0.5\dimexpr \aboverulesep + \belowrulesep + \cmidrulewidth}

\usepackage{color}
\usepackage{iccv}
\usepackage{times}
\usepackage{epsfig}
\usepackage{graphicx}
\usepackage{amsmath}
\usepackage{amssymb}

\usepackage{amsthm}

\theoremstyle{definition}
\newtheorem{definition}{Definition}[section]

\usepackage[breaklinks=true,bookmarks=false]{hyperref}

\iccvfinalcopy 

\def\iccvPaperID{3780} 

\ificcvfinal\fi

\begin{document}

\title{Coarse-to-Fine Searching for Efficient Generative Adversarial Networks} 

\author{Jiahao Wang$^1$, Han Shu$^2$, Weihao Xia$^1$, Yujiu Yang$^1$\thanks{corresponding author}, Yunhe Wang$^2$\\
	$^1$Tsinghua University\\
	$^2$Noah’s Ark Lab, Huawei Technologies\\
	{\tt\footnotesize wang-jh19@mails.tsinghua.edu.cn, xiawh3@outlook.com, yang.yujiu@sz.tsinghua.edu.cn} \\
	{\tt\footnotesize han.shu@huawei.com, wangyunhe@pku.edu.cn}
}

\maketitle
\ificcvfinal\thispagestyle{empty}\fi

\begin{abstract}
This paper studies the neural architecture search (NAS) problem for developing efficient generator networks. 
Compared with deep models for visual recognition tasks, generative adversarial network (GAN) are usually designed to conduct various complex image generation.
We first discover an intact search space of generator networks including three dimensionalities, \ie, path, operator, channel for fully excavating the network performance.
To reduce the huge search cost, we explore a coarse-to-fine search strategy which divides the overall search process into three sub-optimization problems accordingly. 
In addition, a fair supernet training approach is utilized to ensure that all sub-networks can be updated fairly and stably. 
Experiments results on benchmarks show that we can provide generator networks with better image quality and lower computational costs over the state-of-the-art methods. For example, with our method, it takes only about 8 GPU hours on the entire edges-to-shoes dataset to get a 2.56 MB model with a 24.13 FID score and 10 GPU hours on the entire Urban100 dataset to get a 1.49 MB model with a 24.94 PSNR score. 
\end{abstract}

\section{Introduction}
Generative Adversarial Network (GAN)~\cite{goodfellow2014generative} is a kind of neural network with a generator and a discriminator, which has been successfully demonstrated for a series of computer vision tasks including image translation~\cite{zhu2017unpaired, isola2017image, chen2020distilling, yi2017dualgan}, image editing~\cite{biggan, yang2019controllable, jiang2021enlightengan}, image synthesis~\cite{sajjadi2017enhancenet, radford2016unsupervised, gulrajani2017improved, miyato2018spectral}, \etc. The strong visual generation ability of GANs has prompted the industry to deploy them on real-world applications such as smartphones and mobile cameras. Computation cost, model size, and latency should be considered due to the limited hardware resources of mainstream edge devices~\cite{Cai2020Once-for-All}. However, most of the widely used GANs requires enormous computational resources. For example, CycleGAN~\cite{zhu2017unpaired} requires more than 57.3 GFLOPs for processing a  $256\times 256$ image, which is very hard to efficiently applied on edge devices.
\begin{figure*}[t]
\centering
\begin{subfigure}[b]{1\textwidth}
	\includegraphics[width=1.0\linewidth]{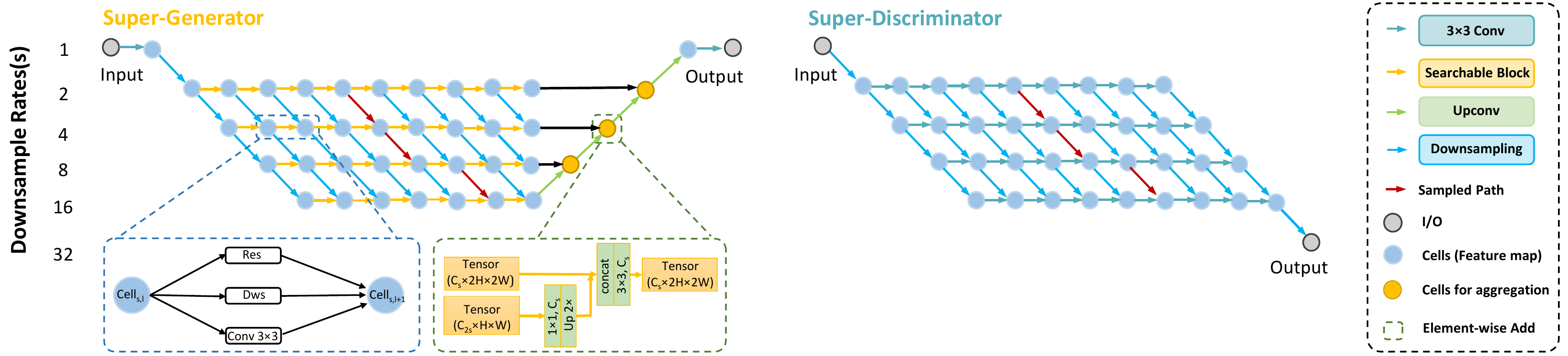}
	\caption{Search Space for Image Translation}
	\label{fig:itsuper}
\end{subfigure}
\begin{subfigure}[b]{1\textwidth}
    \includegraphics[width=1.0\linewidth]{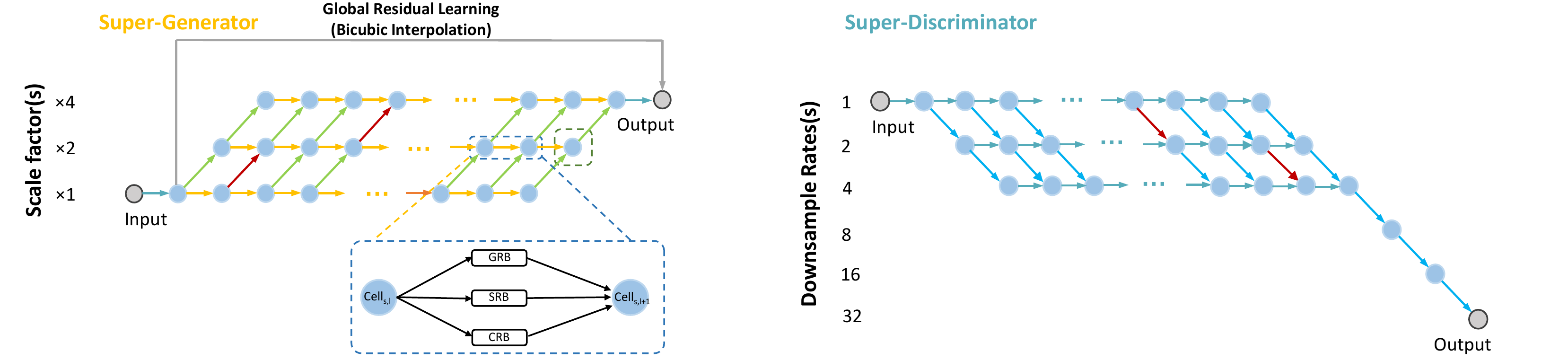}
	\caption{Search Space for Super Resolution}
	\label{fig:srsuper}
\end{subfigure}
\caption{Our hierarchical search space. \textit{Top}: Super-GAN architecture for image translation task. \textit{Bottom}: Super-GAN architecture for super resolution task. These outputs are progressively aggregated together in the end module. For all tasks, sub-generator and the corresponding sub-discriminator are respectively sampled from the super-generator and the super-discriminator (the {\color{red}{red arrow}} shows the sampled path) and we aim to keep the two sub-networks relatively balanced in ability and computationally comparable at the same image resolution. }
\label{fig:pipeline}
\vspace{-10pt}
\end{figure*}

Admittedly, there is already a number of approaches have been developed for compressing and speeding-up deep neural networks. Han~\etal~\cite{han2015deep} first proposed to utilize the weight sparsity to reduce the overall computation overhead of pre-trained deep networks. Wang~\etal~\cite{wang2016cnnpack} further tackled this problem from the perspective of DCT frequency domain to achieve higher compression ratios. However, most of previous model compression methods~\cite{chen2018shallowing,zhuang2018discrimination, he2017channel,zhuang2018discrimination,hinton2015distilling,romero2014fitnets,you2017learning} are designed for recognition networks and suffer from non-negligible performance degradation when directly applied on generator compression. Recently, Shu~\etal~\cite{shu2019co} explored an evolutionary pruning algorithm that can remove redundant filters in pre-defined generator architecture. Chen~\etal~\cite{chen2020distilling} further presented a distillation method to compress both the generator and the discriminator. However, these two approaches cannot discover the structural redundancy within layers and blocks. 

Besides, neural architecture search (NAS~\cite{zoph2018learning, zoph2016neural, lu2018nsga, real2019regularized, liu2018darts, chen2019progressive, cai2018proxylessnas, xu2019pc, li2020sgas, wu2019fbnet, chu2019fair}) is the recent hotspot for automatically finding the best architecture from the given search space. For CNN architectures, Evolutionary Algorithm (EA) based methods~\cite{real2019regularized} maintain a set of architectures and generate new architectures using genetic operations like mutation and crossover. The differentiable NAS methods~\cite{liu2018darts, chen2019progressive, cai2018proxylessnas, li2020sgas, wu2019fbnet} jointly optimize the shared weights and architecture parameters, significantly reducing the search cost and making the search process efficient.
For GAN architectures, Gong~\etal~\cite{2019AutoGAN} proposed the first framework mingling NAS into GANs for the task of unconditional image generation from random noise. Li~\etal~\cite{gancomp_2020} searched the channel width of the ``once-for-all"~\cite{Cai2020Once-for-All} generator and Fu~\etal~\cite{fu2020autogan} performs differentiable neural architecture search~\cite{liu2018darts,chen2019progressive,cai2018proxylessnas,xu2019pc,li2020sgas,wu2019fbnet,chu2019fair} on each block's width and operator. Unfortunately, the mentioned methods have insufficient search space and granularity to fully probe the redundancy and exploit the potential of the network. The architecture of generator can be regarded as a two-level hierarchy, which we defined as outer network level and inner block level. The former is the network path of blocks which controls the spatial resolution and the latter is operators inside blocks which governs specific computation. The commonality of current AutoML based methods~\cite{2019AutoGAN, gancomp_2020, fu2020autogan, wang2020gan} are that they only focus on inner block search while leaving the path level hand-designed. This limited search space becomes problematic for pixel-level task where outer network level variations are sensitive to output image's quality. 
       
To this end,  we present CF-GAN, which extend the search space into three dimensions, \ie path, operator and channel to fully excavating the efficient network structure.  To reduce the intensive computation cost of enlarging the search space,  a Coarse-to-Fine search strategy is proposed and three dimensions of search space is explored sequentially.  Besides,  a fair supernet training approach is utilized to ensure the equal training and evaluation of each sub-generator in different search stage.  Specifically, the fairness reflects in two aspects.  Firstly,  for path level,  a compatible sampling between generator and discriminator is employed. For a certain sampled generator,  the discriminator path is sampled accordingly to keep the two networks maintains comparable capacity on same feature resolution. Thus, each sub-path of generator is trained fairly with a collaborative discriminator with more stability in minimax optimization process.  Secondly,  for operator level,  all the operator candidates are updated at once for several accumulated iterations,  so each operator with same network depth is updated numerically equally. The proposed fair training strategy ensures the fairness of training and evaluation, thus leading to a better performance of the searched architecture. The proposed method is deployed on several benchmarks of image translation and super resolution tasks. The experimental results demonstrate the advantages of model’s computation cost and generated image quality compared with the state-of-the-art methods of GAN compression and efficient searching options.~\cite{gancomp_2020,fu2020autogan,chen2020distilling,shu2019co,ledig2017photo,wang2018esrgan}. 

The rest of this paper is organized as follows: Section~\ref{sec:related_work} investigates
related works on AutoML and GAN. Section~\ref{sec:method} proposes hierarchical architecture search space and the fair super-GAN training strategy. Section~\ref{sec:experiments}
evaluates the proposed method on various benchmark datasets and models and Section~\ref{sec:conclusion} concludes this paper.

\section{Related Work}
\label{sec:related_work}

\paragraph{AutoML and AutoML-Oriented Deep Compression.} AutoML, or Neural Architecture Search (NAS), aims to discover efficient architectures with competitive performance. Reinforcement learning (RL) and genetic algorithms were widely adopted in NAS~\cite{zoph2018learning, zoph2016neural, lu2018nsga, real2019regularized, tan2019mnasnet, 2019AutoGAN}. However, these search methods suffer from massive computational resources and energy consumption. Differentiable architecture search methods~\cite{liu2018darts,chen2019progressive,cai2018proxylessnas,xu2019pc,chu2019fair} propose to address the issue by relaxing the discrete search space and transform the searching task to a joint optimization of architecture parameters and weights via gradient descent. They drastically reduces the searching cost to several days. Even so, they bring both ”Matthew effect” issues, where real potential structures are prematurely excluded, affecting search results. One-shot NAS methods~\cite{guo2019single,bender2018understanding,chu2019fairnas} alleviate the challenge by defining a supernet and decoupling the supernet training and architecture search in two sequential steps. Thus, the trained supernet is used as a performance predictor of architectures to select the most promising architecture.

AutoML has been demonstrated by ~\cite{he2018amc, liu2018demand, wang2019haq, wang2020apq} as an effective measure for deep compression. With an accuracy predictor or an agent controlled by reinforcement learning, the compression ratio can be automatically determined. To our best knowledge, all the above methods focus on compressing deep classifiers, yet at its infancy stage to be employed on GANs. In comparison, the proposed framework provides an innovative idea for latest model compression techniques and is well designed for GAN compression. 

\paragraph{GAN and GAN Compression.}
GANs~\cite{goodfellow2014generative, gui2020review} can be generalized as a non-saturating game optimization which target to minimize different categories of divergences. They can be extended to a conditional version if conditioned on some extra information~\cite{pix2pix2017,mirza2014conditional,gauthier2014conditional}.

Despite GANs have embraced considerable success on numerous tasks such as image translation and super resolution ~\cite{mirza2014conditional,gauthier2014conditional,denton2015deep, pix2pix2017, sangkloy2017scribbler, reed2016generative, zhang2017stackgan, chen2018cartoongan, wang2018pix2pixHD, park2019semantic, Chen_2018_CVPR, taigman2017unsupervised,shrivastava2017learning,zhu2017unpaired,kim2017learning, yi2017dualgan,liu2017unsupervised,choi2017stargan,huang2018multimodal,lee2018diverse,ledig2017photo,wang2018esrgan, guan2019srdgan, yuan2018unsupervised}, the huge computational cost of GANs creates obstacles to mobile deployments, calling for compression techniques. Shu~\etal~\cite{shu2019co} proposed the first algorithm to remove redundant filters in pre-trained CycleGAN, that relied on the cycle-consistency loss and co-evolutionary approach. However, the algorithm was only suitable for compressing CycleGAN or its variants and cannot be extend to encoder-decoder GANs. %
Knowledge distillation techniques are also introduced in GAN compression task~\cite{chen2020distilling,gancomp_2020}. Chen~\etal~\cite{chen2020distilling} jointly distilled generator and discriminator in an adversarial learning process. Li~\etal~\cite{gancomp_2020} further searched channel width of the ``once-for-all" ~\cite{Cai2020Once-for-All} generator. Wang~\etal~\cite{wang2020gan} combined pruning and quantization strategy to establish a unified compression method. Chen~\etal~\cite{ganlottory_2021} compressed GANs by the lottery ticket hypothesis. Previous works usually aim at hand-designing the outer network level of the generator and discriminator architecture, while we are focusing on searching for GAN structures in an automated manner. 

The latest concurrent work~\cite{fu2020autogan} performs differentiable neural architecture search on GANs under the target compression ratio. However, this method is hard to get rid of the ``Matthew effect" issue, which tends to find some blocks with high performance only in the initial phase, resulting in a local minimum solution~\cite{guo2019single,bender2018understanding,chu2019fairnas}. The proposed method could be considered as an important step over the previous method~\cite{fu2020autogan}, that decouples the optimization of architecture parameters and weights with a coarse-to fine search method to diminish the search cost.

\section{Method}
\label{sec:method}

Following the weight sharing NAS method, we decouple the super-GAN training and architecture search in two sequential steps. The architecture search space $G_{\text{super}}$ and $D_{\text{super}}$ are encoded in \textit{supernets}, denoted as $\mathcal{N}(G_{\text{super}}, W_{G})$ and $\mathcal{N}(D_{\text{super}}, W_{D})$, where $W_{G}$ and $W_{D}$ are the weights in the super-generator and super-discriminator, respectively. The supernet is trained on the training set $D_{tr}$ and the optimization target of the supernet can be formulated as:
\vspace{-5pt}
\begin{equation}
\begin{aligned}
   & W_{G_{\text{super}}}, W_{D_{\text{super}}} =
    \mbox{arg} \min_{W_G} \max_{W_D}\\
   \mathcal{L}_\text{train}&\left( \mathcal{N}(G_{\text{super}}, W_{G}), \mathcal{N}(D_{\text{super}}, W_{D}) ; D_{tr}\right).
\end{aligned}
\label{equ:1}
\end{equation}

Then, sub-architectures inherit weights from the supernet. The network structure search is contacted with the validation set $D_{val}$ and formulated as:
\vspace{-5pt}
\begin{equation}
G^* = \mbox{arg}\max_{G\in G_{\text{super}}}\mathcal{H}_\text{val} \left( \mathcal{N}(G, W_{G_{\text{super}}});D_{val}\right).
\label{equ:2}
\end{equation}
where $\mathcal{H}$ is the evaluation metric suitable of different tasks.

\subsection{Hierarchical Architecture Search Space} 
\label{subsec:method_searchspace}
We describe our task-specific hierarchical architecture search space in this section. For each task, we design a super-generator and a super-discriminator, both containing the inner block level and the outer network level. For the inner block level, we inherit the design in~\cite{fu2020autogan, song2019efficient, zoph2018learning, liu2018darts} to keep consistent with pioneer works. For the outer network level, we propose a novel multi-scale search space in which each path is a sequence structure of searchable blocks, and each searchable block represents a single operator to facilitate implementation on the edge-side. 

\paragraph{Network Level Search Space Design.} 
For the super-generator, we leverage the experience from the state-of-the-art generator architectures when designing the task-specific supernet architectures of super resolution and image translation. For image translation task, the spatial size of the next layer is either twice as large or remains the same. We prefer a multi-scale search space that includes several proved efficient blocks~\cite{zhu2017unpaired,pix2pix2017} (See Fig.~\ref{fig:itsuper}). For super-resolution task, the spatial size of the next layer is either twice as small or remains the same. We focus on pixel-wise prediction so that a progressive upsampling~\cite{2017Fast,2018A} search space with the global residual operation is employed (See Fig.~\ref{fig:srsuper}). 

For the super-discriminator, we develop a supernet that is similar in structure to the super-generator, which also includes a multi-scale search space. We aim to keep each sub-discriminator and sub-generator relatively balanced in ability and computationally comparable at the same image resolution. Besides, the output of all sub-discriminators are downsampled to the same resolution.
\begin{figure}[t]
\centering
\vspace{5pt}
\includegraphics[width=1.0\linewidth]{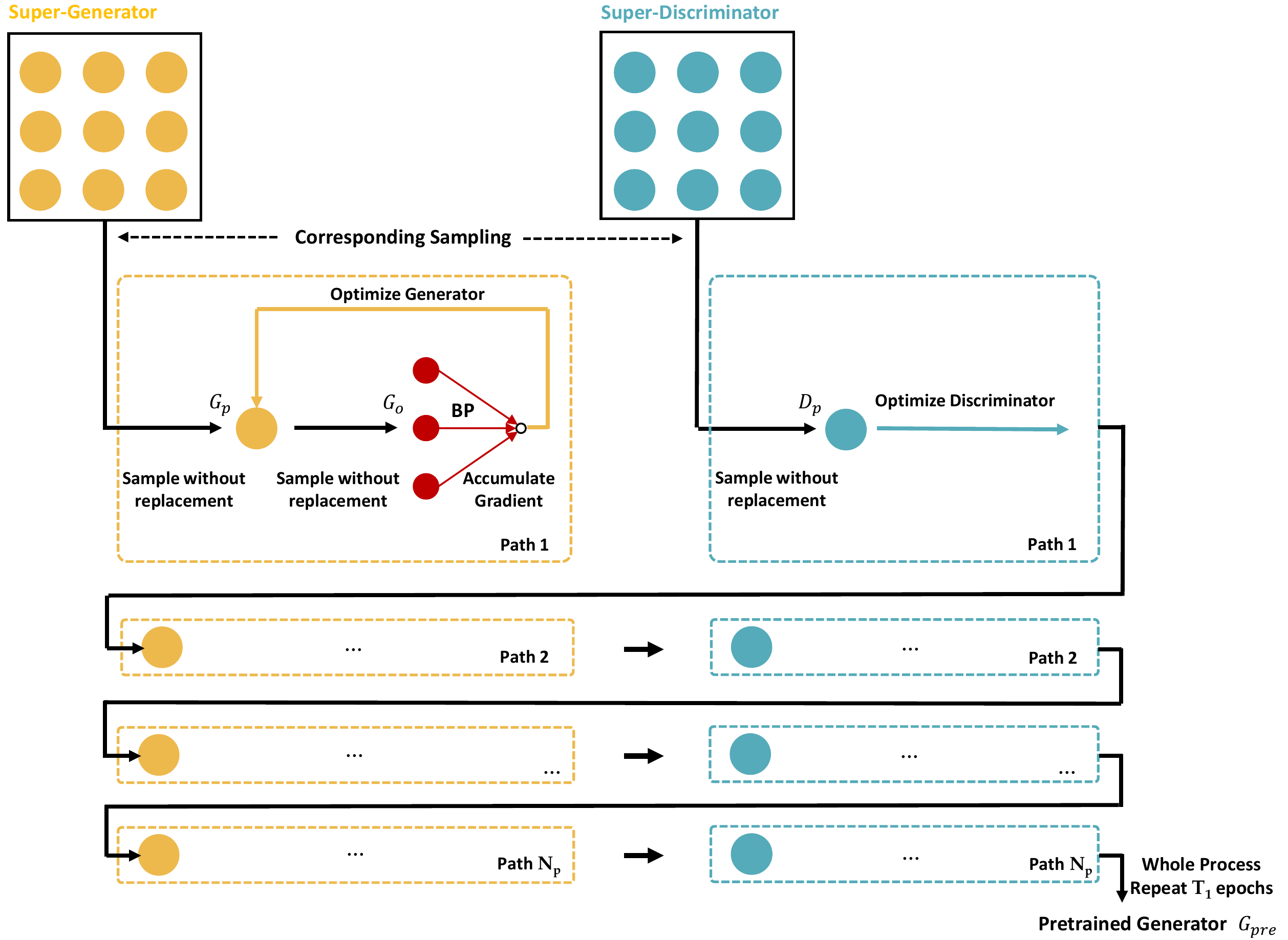}

\caption{
Diagram of the proposed fair super-GAN training strategy. For a certain sampled generator ({\color{yellow}{yellow}} dots), a discriminator path ({\color{teal}{cyan}} dots) is sampled accordingly. All the operator candidates ({\color{red}{red}} dots) from the sampled generator are updated at once. }
\vspace{-10pt}
\label{fig:pretrainG}
\end{figure}

\paragraph{Block Level Search Space Design.} For super-generator, we define a \textit{block} to be an inner searchable unit in each path of the super-generator. 
For image translation, we search for the following operators: Conv 3$\times$3, Residual Block (``ResBlock''), Depthwise Block (``DwsBlock'') to focus on global style and continuity. For super-resolution, we follow~\cite{song2019efficient} to take GRB (Group Residual Block), SRB (Shrink Residual Block), CRB (Contextual Residual Block) as candidate operators. In addition, we enable each layer’s width and operators
to be independently searchable to obtain a crucial balance between the capacity of an operator and its width. Thus, we merge the pruning step into the searching process to build a slimmer generator end-to-end. 

For super-discriminator, the purpose of it is to match a corresponding sub-discriminator for each sub-generator to aid it in finishing stable and thorough training so that only searching in the outer network level will be enough. Specifically, all operators of it are convolution layers with different down-sampling rate following the previous works~\cite{gancomp_2020, fu2020autogan}. 

\begin{figure}[t]
\centering
\begin{minipage}{0.45\textwidth}
\vspace{-1em}
\begin{algorithm}[H]
\caption{Fairly pretrain the supernet generator}
\label{alg:pretrain}
\begin{algorithmic}[1]
\REQUIRE{$D_{tr}$, $G_{\text{super}}$, $D_{\text{super}}$, $G_0$, $N_{p}$, $M$, $T_1$, $\left\{\alpha^{(t)}\right\}$, $\left\{\eta^{(t)}\right\}$.}\\
\STATE Initialize the super-GAN's weight $\{ w_G,w_D \}$ and architecture parameters ${\gamma}$ with uniform distribution. 
\FOR{$t$ $\gets$ 1\ to\ $T_1$ } 
\STATE Get data $(X,Y)$ from the training set $D_{tr}$. 
\FOR{$p$ $\gets$ 1\ to\ $N_p$ } 
\STATE Uniform sample $G_p$ from $G_{\text{super}}$ and corresponding $D_p$ from $D_{\text{super}}$ w/o replacement.
\STATE Clear gradients recorder for all weights.
\FOR{$m$ $\gets$ 1\ to\ $M$}
\STATE Sample $G_o$ from $G_p$ w/o replacement.
\STATE $g_{G_o(w_G)}^{(t)} = \nabla_{w_{G}} L_{\text{train}}(G_o,G_0,D_p,X,Y)$
\STATE Accumulate gradients $g_{G_o(w_G)}^{(t)}$ to $g_{G_p(w_G)}^{(t)}$.
\ENDFOR
\STATE $w_{G}^{(t+1)} \gets w_{G}^{(t)} - \alpha^{(t)} g_{G_p(w_{G})}^{(t)}$
\STATE $g_{G_p(\gamma)}^{(t)} = \nabla_\gamma L_{\text{train}}(G_p,G_0,D_p,X,Y)$
\STATE $\gamma^{(t+1)} \gets \textrm{prox}_{\lambda_{sp}\eta^{(t)}} (\gamma^{(t)} - \eta^{(t)} g_{G_p(\gamma)}^{(t)})$
\STATE $g_{D_p(w_D)}^{(t)} = \nabla_{w_{D}} L_{\text{train}}(G_p,G_0,D_p,X,Y)$
\STATE $w_{D}^{(t+1)}$ $\gets$ $w_{D}^{(t)} + \alpha^{(t)} g_{D_p(w_D)}^{(t)}$
\ENDFOR
\ENDFOR
\ENSURE{Pre-trained super-generator $G_{\text{pre}}$.}
\end{algorithmic}
\end{algorithm}
\end{minipage}
\vspace{-1.25em}
\end{figure}

\subsection{Fair Super-GAN Training Strategy} 
Training super-GAN matters since it serves as a fundamental performance predictor for different sub-networks. Due to the consideration of memory consumption, previous One-Shot NAS methods~\cite{guo2019single, bender2018understanding, chu2019fairnas} implement training by sampling a single route from the supernet. We formulate this approach in our special GAN architecture as:
\begin{equation}
\begin{aligned}
    W_{G_{\text{super}}}, W_{D_{\text{super}}} =
    \mbox{arg} \min_{W_G} \max_{W_D}
\mathbb{E}_{G \sim p(G_{\text{super}}),D \sim q(D_{\text{super}})}\\\left[
   \mathcal{L}_\text{train}\left( \mathcal{N}(G, W_{G}), \mathcal{N}(G, W_{D}) ; D_{tr}\right)\right].
\end{aligned}
\label{equ:3}
\end{equation}

where $p$ and $q$ are discrete sampling distribution over $G_{\text{super}}$ and $D_{\text{super}}$, which is usually preset to be uniform sampling~\cite{guo2019single, bender2018understanding}. However, this measure will lead to a huge gap between the performances by prediction of the supernet and by training from scratch. We attribute this drawback to the unfair training of all sub-networks, which fails to fully exploit the potential of each sub-network. In addition, the GAN training is notoriously unstable, making this problem more serious on super-GAN training. So how to build a good evaluator that can accurately estimates each sub-generator's performance? We answer the question by proposing an fair training strategy for GAN architecture. 

\begin{figure}[t]
\centering
\begin{minipage}{0.46\textwidth}
\vspace{-1em}
\begin{algorithm}[H]
  \caption{The proposed CF-GAN framework}
  \label{alg:GSL}
    \begin{algorithmic}[1]
        \REQUIRE{$D_{tr}$, $D_{val}$, $G_{\text{super}}$, $D_{\text{super}}$, $G_0$, $N_{p}$, $N_{o}$}\\
        \textit{\# First Step: Fairly Pretrain the Supernet Generator}\\
        \STATE Obtain the pre-trained $G_{\text{super}}$ by Alg~\ref{alg:pretrain} as $G_{\text{pre}}$.\\
        \textit{\# Second Step: Coarse-to-fine Search}
        \FOR{$n$ $\gets$ 1 \ to\ $N_p$ }
        \STATE Sampling subnet $G_n$ from $G_{\text{pre}}$ w/o replacement.
        \STATE Inherit weights of $G_n$ directly from $G_{\text{pre}}$. 
        \STATE $f_n$ = Evaluate($G_n$, $D_{val}$)
        \ENDFOR
        \STATE Select suitable architecture: $G_{\text{path}}={\underset {G_n}{{arg\,max} }}\,(f_n)$
        \FOR{$m$ $\gets$ 1 \ to\ $N_o$ }
        \STATE Sampling subnet $G_m$ from $G_{\text{path}}$ w/o replacement. 
        \STATE Inherit weights of $G_m$ directly from $G_{\text{path}}$. 
        \STATE $f_m$ = Evaluate($G_m$, $D_{val}$)
        \ENDFOR
        \STATE Select suitable architecture: $G_{\text{optr}}={\underset {G_m}{{arg\,max} }}\,(f_m)$\\
        \STATE Obtain architecture $G_{\text{channel}}$ from $G_{\text{optr}}$ by shrinking redundant channels with evolutionary algorithm.\\
        \textit{\# Third Step: Fine-tune}
        \STATE Fine-tune $G_{\text{channel}}$ to obtain the output $G^*$.
        \ENSURE{The compressed efficient generator $G^*$.}\\
\end{algorithmic}
\end{algorithm} 
\end{minipage}
\vspace{-1em}
\end{figure}

\paragraph{Total Training Objective.} 
We use a combination of GAN loss $\mathcal{L}_\text{GAN}$~\cite{goodfellow2014generative}, perceptual loss $\mathcal{L}_\text{per}$~\cite{2016Perceptual} to measure high level differences, reconstruction loss $\mathcal{L}_\text{recon}$~\cite{gancomp_2020} to unify the unpaired and paired learning. In addition, we take the sparsity regularization $\mathcal{L}_\text{sp}$~\cite{liu2017learning} to automatically identify the redundant channels to prune. Specifically, $L_1$ norm on the trainable scaling factor $\gamma$ is applied in the normalization layers to encourage channel sparsity. The objective function can be summarized as:
\begin{equation}
    \label{eqn:loss}
    \mathcal{L_{\text{train}}} =   \mathcal{L}_\text{GAN} + \lambda_{\text{recon}} \mathcal{L}_\text{recon} + \lambda_{\text{per}} \mathcal{L}_\text{per}+ \lambda_{\text{sp}} \mathcal{L}_\text{sp},
\end{equation}
where $\lambda_{\text{recon}}$, $\lambda_{\text{per}}$, $\lambda_{\text{sp}}$ are hyper-parameters.

\paragraph{Fair Training Strategy for GAN architecture.}
Considering the three search dimensionalities of the generator architecture (\ie path, operator, channel), we discuss the fairness of super-GAN training from two views.\\
\textit{Path View}: Each path is trained fairly. We regard the multi-scale search space as a set of different paths, each representing a sub-generator. We guarantee this by matching each sub-generator with a specific sub-discriminator with comparable capabilities for stable training and fully exploiting the potential of each path.
\\ 
\textit{Operator View}: Different operators in each path are fairly trained. In a certain layer of a certain path of the super-generator, the amount of times the parameters of each operator be updated is numerically equal at all times. 

\begin{definition}[\textbf{Fair Super-GAN Training}] Given one of the totally $N_p$ paths from the super-generator that consists of $L$ layers, each with $M$ candidate choice blocks. Suppose the weights are totally updated for $T$ times. Denote $U_{p,l_{m}}^{t}$ as the times of updates over the first $t$ trials.
$m$,$l$,$p$ are the index of operation, layer, and path respectively. Denote $U_{p}^{t}$, $V_{p}^{t}$ as the number of times the $p$-th path of the super-generator and the  super-discriminator is updated over the first $t$ trials, respectively. $\forall l\in [1,L]$, $t\in [1,T]$, $\forall p\in [1,N_p]$:
\vspace{2pt}
\begin{itemize}[leftmargin=*]
    \item \textit{Path View}: $U_{p}^{t}=V_{p}^{t}$ holds.
	\vspace{5pt}
	\item \textit{Operator View}: $U_{p,l_{1}}^{t}=U_{p,l_{2}}^{t}=\hdots=U_{p,l_{M}}^{t}$ holds. 
\end{itemize}
\end{definition}
\vspace{-2pt}

In the One-Shot NAS method for classification tasks, one method for training supernet is to uniformly sample one operation at a time and update its weight parameters. Each sampling process is independent, and the number of times each operation is sampled during training follows a binomial distribution. We show that such measure cannot meet the \textit{Fair Super-GAN Trainin} in the GAN model, as applied in Appendix A.1. 

To this end, we propose a novel training strategy that strictly meets the \textit{Fair Super-GAN Training} definition, as shown in Fig.~\ref{fig:pretrainG}. For each step, we uniformly sample one path from $G_{\text{super}}$ without replacement dubbed $G_{p}$ to optimize. A discriminator path $D_{p}$ is sampled accordingly from $D_{\text{super}}$ to keep the two networks maintaining comparable capacity on same feature resolution. Thus, each sub-path of generator is trained fairly with a collaborative discriminator with more stability following a minimax two-player game. Specifically for each path $G_p$'s training process, we sample $M$ sub-generator without replacement from $G_p$ to build $M$ subnets. We do not immediately perform back-propagation and update parameters to reduce the bias of training orders. Instead, we calculate the gradients of the $M$ models and accumulate them. After completing all $M$ BPs, we update the weights of $G_p$ with the accumulated gradients to complete the optimization process of $G_p$ in the current epoch. Then, we update the parameters of discriminator $D_{p}$ according to the minimax optimization of GANs and complete the on-going cycle. Then we go to the next cycle and repeat sampling a new $G_p$ without replacement. On completing $N_p$ cycles, the parameters of each path in $G_{\text{super}}$ and $D_{\text{super}}$ are updated once, marking the end of an epoch. The whole process is repeated $T_1$ epochs and the output model is the pre-trained $G_{\text{super}}$ dubbed $G_{\text{pre}}$, as described in Alg.~\ref{alg:pretrain}. The detailed version of Alg.~\ref{alg:pretrain} is shown in Appendix A.3.

\subsection{Coarse-to-fine Search Strategy} 
\label{sec:cf}
To speed up the search procedure and reduce massive computation cost, we propose to conduct a searching order from large sub-generators to small ones in a coarse-to-fine manner (See Alg.~\ref{alg:GSL}, the detailed version of Alg.~\ref{alg:GSL} is shown in Appendix A.4.), which is detailed as follows:
\begin{itemize}[leftmargin=*]
\item \textbf{Auto Path Design}. 
To find the optimal path, each of the $N_p$ different paths is activated in turn with weights inherited from the supernet. Each path is composed of sequential searchable blocks with mixed operations. We only conduct an inference process on the path and select one with highest evaluation score as the optimal path design.

\item \textbf{Auto Operator Specialization}. Given the optimal path, the next goal is to find the optimal operator specialization that is from a total of $N_o = (M!)^{L-1}$ (the proof is in Appendix A.2) choices. We activate a candidate operation in each layer in turn to build a subnet. Similar to the previous step, we evaluate all subnets and select one with highest evaluation score as the optimal operator specialization. 

\item \textbf{Auto Channel Shrinking}. 
We give each block the flexibility to choose different channel expansion ratios, then the redundant channels are automatically identified and pruned. We employ a channel configuration set $\{c_1, c_2, ..., c_i, ..., c_K\}$, where $K$ indicates the granularity of pruning. The number of possible configurations increases exponentially as $K$ increases, resulting in a far too time-consuming search process. Therefore, we resort to the evolutionary algorithm with the directional mutation\footnotemark to guide the evolution process and find the best channel that satisfies diverse computational constraints. 
\footnotetext{Detailed evolutionary algorithm is shown in Appendix A.5.}
\end{itemize}
\vspace{3pt}

The computational complexity of the joint search method $O_{\text{Joint}}$ is the product of that ones of the three dimensions:
\vspace{-5pt}
\begin{equation}
\small
O_{\text{Joint}}=O_{\text{Path}}\cdot O_{\text{Operator}}\cdot O_{\text{Channel}}
\label{cost joint}
\vspace{-5pt}
\end{equation}

With this pipeline, the computational complexity of the search $O_{\text{Coarse-to-fine}}$ shifts from the product to the sum of that ones of the three dimensions, which can be formulated as:

\vspace{-10pt}
\begin{equation}
\small
O_{\text{Coarse-to-fine}}=O_{\text{Path}}+ O_{\text{Operator}}+ O_{\text{Channel}}\ll O_{\text{Joint}}
\label{cost GSL}
\end{equation}

In practice, we could save up to 90\% search time of joint search pipeline with on par performance, as shown in Tab.~\ref{tab:searchcost}.

\section{Experiments}
\label{sec:experiments}
\subsection{Experiment Setting} 
\label{sec:Setups}

\paragraph{Tasks and Datasets.} We apply the proposed method on two representative GAN based tasks, image translation and super resolution. For image translation, we consider model Pix2Pix~\cite{pix2pix2017} and CycleGAN ~\cite{zhu2017unpaired}, respectively on their benchmarks: edges$\to$shoes~\cite{yu2014fine}, Cityscapes~\cite{cordts2016cityscapes}, map$\to$arial photo and horse$\to$zebra~\cite{zhu2017unpaired}, zebra$\to$horse. For super-resolution, we consider compressing GAN-based SISR model on a combined dataset of DIV2K and Flickr2K~\cite{timofte2017ntire} with a upscale factor of 4$\times$. We evaluate our model on several popular SR benchmarks:  Set5~\cite{bevilacqua2012low}, Set14~\cite{zeyde2010single},  BSD100~\cite{martin2001database} and Urban100~\cite{huang2015single}.

\paragraph{Evaluation Metrics.} For model performance metrics, we use FID~\cite{heusel2017gans} for image translation, and PSNR for super-resolution. For model complexity measurement, we consider the model size and the FLOPs (G). We adopt the memory consumption of the parameters (MB) for model size evaluation following AGD ~\cite{fu2020autogan}.

\begin{figure*}[t]
\centering
\vspace{-15pt}
\includegraphics[width=\linewidth]{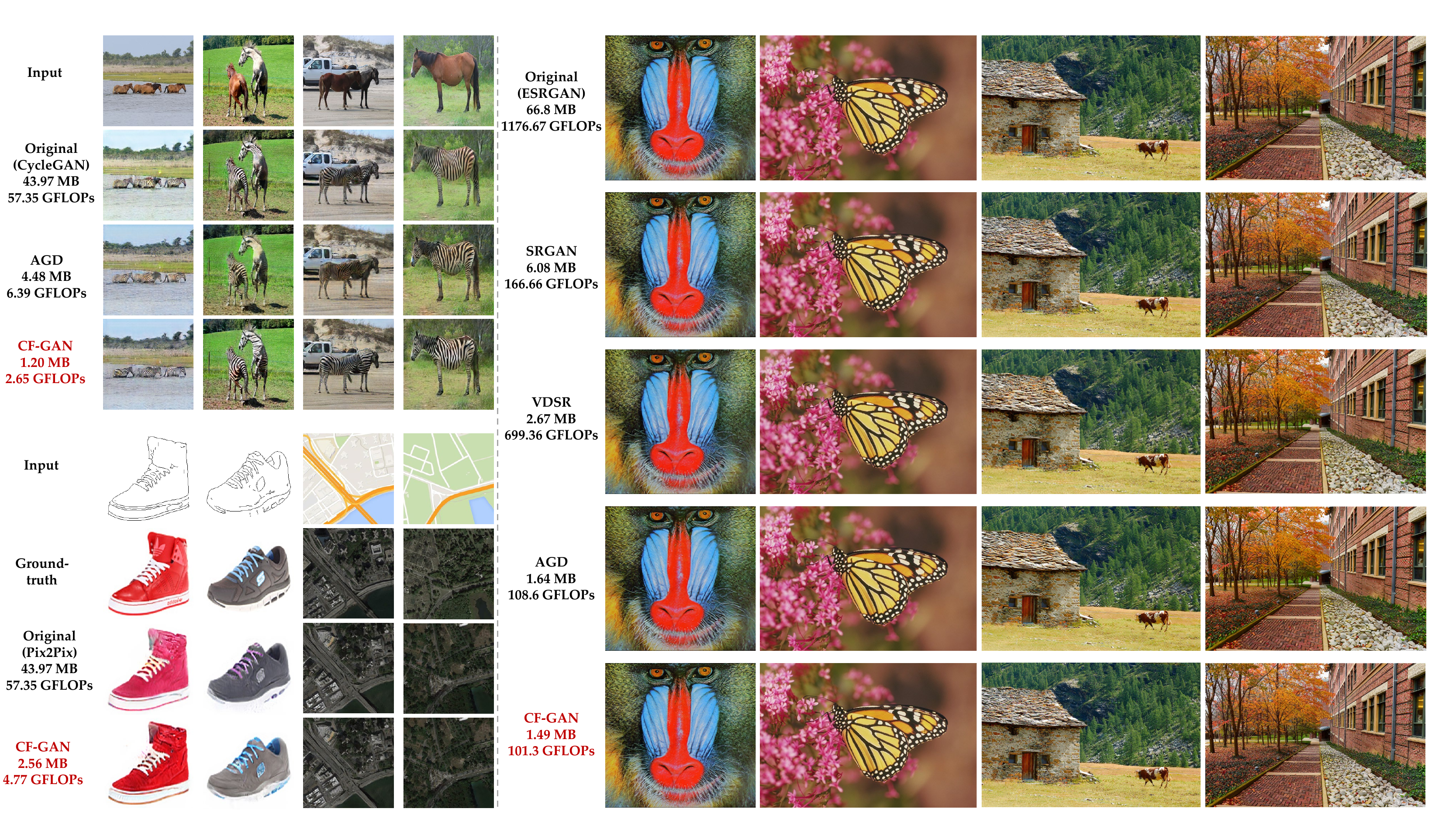}
\caption{Qualitative results on different super resolution methods, CycleGAN and Pix2Pix compression with benchmarks. For all models we compare FLOPs (G) and model size of different compression methods. Our CF-GAN framework highly preserves the fidelity while significantly reducing the model complexity. Considering image quality, our method is better than the recent AGD~\cite{fu2020autogan} strategy, and is far superior to previous methods (SRGAN~\cite{ledig2017photo}, VDSR~\cite{kim2016accurate}) in super-resolution tasks.}
\label{fig:qualitative}
\vspace{-10pt}
\end{figure*}

\subsection{Implementation Details} 
\label{sec:implementation}
For image translation, we pre-train the supernet for 40 epochs to avoid the architecture collapse~\cite{liu2018darts}, with batch size 1. We adopt the Adam optimizer and keep a constant learning rate $2\times 10^{-4}$ in the beginning and linearly decay the rate to zero in the later stage. For super-resolution, we searched the repeated times of a certain block from $\{1, 2, 3, 4\}$, which are defined as recursive number. The recursive number ${1}$ denotes one normal convolution layer. We pre-train the supernet for 100 epochs with a batch of 6 cropped patches with $32\times32$ and an ADAM optimizer with an initial learning rate to be $1\times 10^{-4}$ which halves for every 25 epochs. Hyper-parameters $\lambda_{\text{recon}}$, $\lambda_{\text{per}}$ and $\lambda_{\text{sp}}$ in Eq.~\ref{eqn:loss} are set to 10, 100 and $1\times 10^{-3}$. We fine-tune the searched architecture for 600 epochs, with batch size of 16. We adopt the Adam optimizer and keep a constant learning rate $1.5\times 10^{-4}$ in the beginning and linearly decay the rate to zero in the later stage of the training. 

\subsection{CF-GAN for Image Translation} 
\label{sec:imagetranslation}

\begin{table}[!t]
\renewcommand*{\arraystretch}{1.15}
\setlength{\tabcolsep}{3pt}
\small\centering
\resizebox{0.5\textwidth}{!}{
\begin{tabular}{clccccc}
\toprule
\multirow{2}{*}[\multirowcenter]{Dataset} & \multirow{2}{*}[\multirowcenter]{Method} & \multicolumn{1}{c}{\multirow{2}{*}[\multirowcenter]{GFLOPs}} & \multicolumn{1}{c}{\multirow{2}{*}[\multirowcenter]{Mem. (MB)}}&
\multirow{2}{*}[\multirowcenter]{Lat. (ms)}& \multicolumn{2}{c}{Metric} \\ 
\cmidrule(lr){6-7}
& & & & &\multicolumn{1}{c}{FID $\downarrow$} & \multicolumn{1}{c}{mIoU $\uparrow$} \\
\midrule
  \multirow{6}{*}{edges2shoes} & Original~\cite{pix2pix2017} & 57.35 & 43.97 & 10.03 & 25.16 & \multicolumn{1}{c}{--} \\
&  Pruned & 10.33  & 6.34 & 7.32 & 35.30 & \multicolumn{1}{c}{--}\\
&  GAN Distill.~\cite{chen2020distilling} & 5.15  & 4.96 & 6.56 & 28.31 & \multicolumn{1}{c}{--} \\
&  GAN Comp.~\cite{gancomp_2020} & 4.83  & 2.80 & 4.80 & 26.60 & \multicolumn{1}{c}{--} \\
&  AGD~\cite{fu2020autogan} & 7.96 & 4.82 & 4.94 & 29.65  & \multicolumn{1}{c}{--} \\
&  \textbf{CF-GAN} & \textbf{4.77}  & \textbf{2.56} & \textbf{4.60} & \textbf{24.13} & \multicolumn{1}{c}{--} \\
\cmidrule{1-7}

\multirow{6}{*}{Cityscapes} & Original~\cite{pix2pix2017} & 57.35& 43.97& 11.8&\multicolumn{1}{c}{--}& 42.01    \\
&  Pruned & 8.97  & 6.34 & 7.94 &\multicolumn{1}{c}{--} & 38.84 \\
&  GAN Distill.~\cite{chen2020distilling} & 5.65 & 5.04 & 7.83 &\multicolumn{1}{c}{--} & 41.05 \\
&  GAN Comp.~\cite{gancomp_2020} & 5.73  & 2.84 & 7.32 &\multicolumn{1}{c}{--} & 40.77  \\
& AGD~\cite{fu2020autogan} & 8.20  & 6.45 & 7.48 &\multicolumn{1}{c}{--}& 41.55 \\
&  \textbf{CF-GAN} & \textbf{5.62}  & \textbf{2.60} & \textbf{7.15} &\multicolumn{1}{c}{--} & \textbf{42.24}  \\
\cmidrule{1-7}

  \multirow{6}{*}{map2arial photo} & Original~\cite{pix2pix2017} & 57.35 & 43.97 & 9.58 & 49.92  & \multicolumn{1}{c}{--} \\
&  Pruned & 9.61  & 6.34 & 5.66 & 63.33 & \multicolumn{1}{c}{--}\\
&  GAN Distill.~\cite{chen2020distilling} & 4.98  & 4.93 & 5.48 & 51.24 & \multicolumn{1}{c}{--} \\
&  GAN Comp.~\cite{gancomp_2020} & 4.72  & 3.00 & 5.11 & 46.30  & \multicolumn{1}{c}{--} \\
&  AGD~\cite{fu2020autogan} & 7.64  & 5.22 & 5.32 & 49.97  & \multicolumn{1}{c}{--} \\
& \textbf{CF-GAN} & \textbf{4.50}  & \textbf{2.80} & \textbf{5.05} & \textbf{46.15} & \multicolumn{1}{c}{--} \\
\bottomrule
\end{tabular}
}
\vspace{2pt}
\caption{
Quantitative comparison with the latest methods on Pix2Pix compression in paired image translation tasks.}
\label{tab:finalpix}
\vspace{-10pt}
\end{table}

\paragraph{Quantitative Results.} We report the quantitative results of compressing Pix2Pix and CycleGAN on their benchmarks respectively on Table~\ref{tab:finalpix} and Table~\ref{tab:finalcycle}. For all tasks, CF-GAN achieves state-of-the-art FID with even smaller memory consumption and lower computational cost, compared with the latest AGD~\cite{fu2020autogan} and GAN compression~\cite{gancomp_2020}. Compared with original CycleGAN and Pix2Pix model, models obtained by proposed method are respectively \textbf{26.4$\times$} and \textbf{36.6$\times$} smaller on computational cost and \textbf{12.0$\times$} and \textbf{17.2$\times$} smaller on model size, with even better model performance. 
\begin{table}[t]
\renewcommand*{\arraystretch}{1.15}
\setlength{\tabcolsep}{4.5pt}
\small\centering
\resizebox{1\linewidth}{!}{
\begin{tabular}{clcccc}
\toprule

\multirow{1}{*}[\multirowcenter]{Dataset} & \multirow{1}{*}[\multirowcenter]{Method} & \multicolumn{1}{c}{\multirow{1}{*}[\multirowcenter]{GFLOPs}} & \multicolumn{1}{c}{\multirow{1}{*}[\multirowcenter]{Mem. (MB)}}&
\multirow{1}{*}[\multirowcenter]{Lat. (ms)}& \multicolumn{1}{c}{\multirow{1}{*}[\multirowcenter]{FID$\downarrow$}}\\
\midrule
\multirow{7.5}{*}{horse2zebra} & Original & 57.35  & 43.97  & 7.25 & 74.12  \\
 & Pruned & 4.95 & 3.91 & 6.19 & 220.9 \\
 & GAN Distill.~\cite{chen2020distilling} & 4.66  & 4.13 & 5.88 & 73.28   \\
 & CEC~\cite{shu2019co} & 13.45  & 10.16 & 6.78 & 96.15 \\
 & GAN Comp.~\cite{gancomp_2020} & 2.73 & 1.40 & 4.03 & 64.95  \\
 & GS-8~\cite{wang2020gan} & 10.99  & 2.00 & 6.36 & 88.14 \\
 & AGD~\cite{fu2020autogan} & 6.39 & 4.48 & 4.07 & 83.6 \\
 & \textbf{CF-GAN} & \textbf{2.65} & \textbf{1.20} & \textbf{3.98} & \textbf{62.31}\\

\cmidrule{1-6}
\multirow{7.5}{*}{zebra2horse} & Original & 57.35  & 43.97 & 7.25 & 147.68 \\
 & Pruned & 4.95 & 3.91& 6.19 & 206.56\\
 & GAN Distill.~\cite{chen2020distilling} & 4.66  & 4.66 & 5.88 & 152.67\\
 & CEC~\cite{shu2019co} & 13.06 & 10.00 & 6.78 & 157.9 \\
 & GAN Comp.~\cite{gancomp_2020} & 2.73 & 1.4 & 4.03 & 146.11 \\
 & GS-8~\cite{wang2020gan} & 12.02 & 2.07 & 6.36 & 119.05 \\
 & AGD~\cite{fu2020autogan} & 4.84 & 3.20 & 3.99 & 137.2\\
 & \textbf{CF-GAN} & \textbf{2.65} & \textbf{1.20} & \textbf{3.88} & \textbf{120.15}\\
\bottomrule
\end{tabular}
}
\vspace{2pt}
\caption{
Quantitative comparison with the latest methods on CycleGAN compression in unpaired image translation tasks.}
\label{tab:finalcycle}
\vspace{-10pt}
\end{table}

We also compare the inference latency(ms) on 2080Ti GPU as it is directly related to user experience. Models obtained by our method also achieve the minimum latency for all experiments, which demonstrates the universality and efficiency of CF-GAN.

\paragraph{Qualitative Results.} Fig.~\ref{fig:qualitative} shows the visual quality from the original model (CycleGAN, Pix2Pix), ground-truth (only for Pix2Pix), CF-GAN, and AGD on benchmark datasets. CF-GAN shows better visual quality compared with the competitors. It is worth mentioning that for some samples, the results generated by the original model show chromatic aberration, but our results mitigate chromatic aberration to some extent. 

\subsection{CF-GAN for Super-Resolution} 
\label{sec:superresolution}
\paragraph{Quantitative Results.} We report the quantitative results in Table~\ref{tab:finalsr}. CF-GAN also achieves state-of-the-art PSNR among a series of GAN-based SR models with the largest compression ratio on model size (over 44 times) and computational cost (over 11 times). 

\begin{table}[!t]
\renewcommand*{\arraystretch}{1.15}
\setlength{\tabcolsep}{4.5pt}
\small\centering
\resizebox{1\linewidth}{!}{
\begin{tabular}{clcccc}
\toprule

\multirow{1}{*}[\multirowcenter]{Dataset} & \multirow{1}{*}[\multirowcenter]{Method} & \multicolumn{1}{c}{\multirow{1}{*}[\multirowcenter]{GFLOPs}} & \multicolumn{1}{c}{\multirow{1}{*}[\multirowcenter]{Mem. (MB)}} &\multirow{1}{*}[\multirowcenter]{Lat. (ms)}& \multicolumn{1}{c}{\multirow{1}{*}[\multirowcenter]{PSNR $\uparrow$}}\\
\midrule
\multirow{6}{*}{Set5} & SRGAN~\cite{ledig2017photo} & 166.66 & 6.08 & 4.92 & 29.40\\
& ESRGAN~\cite{wang2018esrgan} & 1176.67 & 66.8 & 84.62 & 30.47 \\
& Pruned & 113.07 & 6.40 & 64.75 & 28.07   \\
& AGD~\cite{fu2020autogan} & 108.6 & 1.64 & 5.08 & 30.44  \\
& \textbf{CF-GAN} & \textbf{101.3}  & \textbf{1.49} & \textbf{4.90} & \textbf{30.62}  \\
\cmidrule{1-6}
\multirow{6}{*}{Set14} & SRGAN~\cite{ledig2017photo} & 166.66 & 6.08 & 5.37 & 26.02\\
& ESRGAN~\cite{wang2018esrgan} & 1176.67 & 66.8 & 80.03 & 26.29 \\
& Pruned & 113.07 & 6.40 & 78.47 & 25.21   \\
& AGD~\cite{fu2020autogan} & 108.6 & 1.64 & 6.91 & 27.28  \\
& \textbf{CF-GAN} & \textbf{101.3}  & \textbf{1.49} & \textbf{5.16} & \textbf{27.94}  \\
\cmidrule{1-6}
 \multirow{6}{*}{BSD100} & SRGAN~\cite{ledig2017photo} & 166.66 & 6.08 & 4.15 & 24.18\\
& ESRGAN~\cite{wang2018esrgan} & 1176.67 & 66.8 & 81.34 & 25.32 \\
& Pruned & 113.07 & 6.40 & 69.43 & 24.74   \\
& AGD~\cite{fu2020autogan} & 108.6 & 1.64 & 5.11 & 26.23  \\
& \textbf{CF-GAN} & \textbf{101.3}  & \textbf{1.49} & \textbf{4.23} & \textbf{26.65}  \\
\cmidrule{1-6}
 \multirow{6}{*}{Urban100} & SRGAN~\cite{ledig2017photo} & 166.66 & 6.08 & 25.28 & 24.39\\
& ESRGAN~\cite{wang2018esrgan} & 1176.67 & 66.8 & 184.79 & 24.36 \\
& Pruned & 113.07 & 6.40 & 113.82 & 22.67   \\
& AGD~\cite{fu2020autogan} & 108.6 & 1.64 & 21.6 & 24.74  \\
& \textbf{CF-GAN} & \textbf{101.3}  & \textbf{1.49} & \textbf{20.53} & \textbf{24.94}  \\
\bottomrule
\end{tabular}
}
\vspace{2pt}
\caption{
Quantitative comparison with the latest methods on ESRGAN compression in super resolution tasks.}
\label{tab:finalsr}
\vspace{-15pt}
\end{table}

\begin{figure*}[h]
\centering
\vspace{-5pt}
\includegraphics[width=1.0\linewidth]{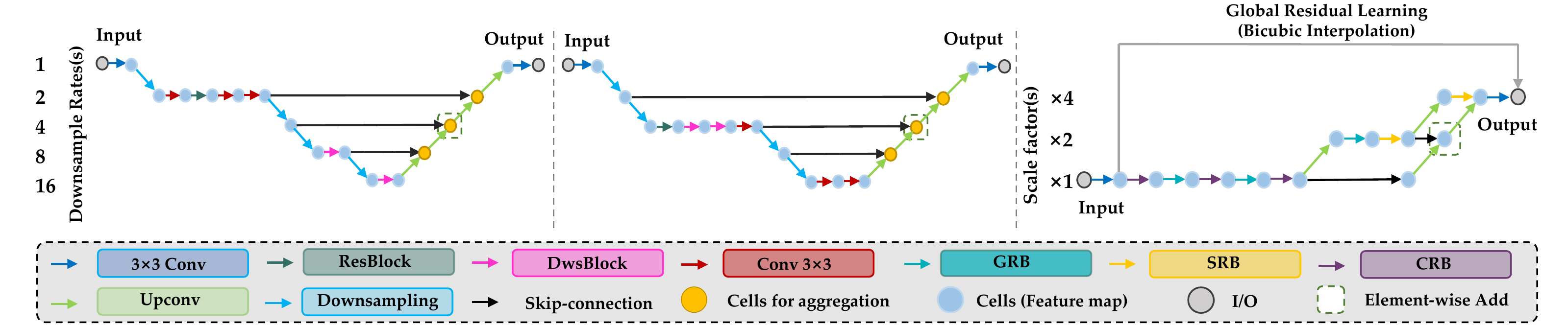}
\caption{Searched architecture by our proposed CF-GAN on Pix2Pix (Left), CycleGAN (Middle) and ESRGAN (Right).}

\vspace{-5pt}
\label{fig:searched_all_h}
\end{figure*}
\paragraph{Qualitative Results.} As shown in Fig.\ref{fig:qualitative} and Appendix B.2, CF-GAN outperforms other methods on producing color richness, sharp edges, and texture details, confirming the effectiveness of the NAS method applied to the GAN Model. 

\subsection{Discussion of the Searched Architecture} 
\label{sec:inspiration}

The searched architectures for all tasks are shown in Fig.~\ref{fig:searched_all_h} and Appendix B.1. The detailed recursive numbers and channel configuration table is provided in the supplement.
We obtain some useful inspirations by analyzing the searched architecture, which can be used as an important guide for the subsequent design of related model architectures.

\begin{table}[t]
\renewcommand*{\arraystretch}{1.25}
\setlength{\tabcolsep}{4.5pt}
\small\centering
\resizebox{1\linewidth}{!}{
\begin{tabular}{lccccc|cc}
\toprule
\multirow{1}{*}[\multirowcenter]{Method} & \multirow{1}{*}[\multirowcenter]{Path} & \multicolumn{1}{c}{\multirow{1}{*}[\multirowcenter]{Operator}} & \multicolumn{1}{c}{\multirow{1}{*}[\multirowcenter]{Channel}}&
\multirow{1}{*}[\multirowcenter]{Fair}&
\multirow{1}{*}[\multirowcenter]{Coarse-to-fine}&
\textsc{FID ($\downarrow$)}&
\textsc{GFLOPs}\\
\midrule
Distill.~\cite{chen2020distilling}& A& A& A& & &73.28&4.66  \\
CEC~\cite{shu2019co}& A& A& A& & &96.15&13.45\\
GS-32~\cite{wang2020gan}& A& A& A& & &86.09&11.35\\
GS-8~\cite{wang2020gan}& A& A& A& & &88.14&10.99\\
AO& N& A& N&\checkmark &\checkmark &76.78&6.35\\
AP& A& N& N&\checkmark &\checkmark &84.35&10.67\\
w/o Fair Training& N& N& N& &\checkmark &65.37&4.36\\
Joint Search& N& N& N&\checkmark &  &62.30&4.43\\
GAN Comp.~\cite{gancomp_2020}& A& A& N& & &64.95 &2.73\\
AGD~\cite{fu2020autogan}& A& N& N& & &83.6 &6.39\\
\textbf{CF-GAN}& N& N& N&\checkmark &\checkmark &\textbf{62.31} &\textbf{2.65}\\
\bottomrule
\end{tabular}
}
\vspace{2pt}
\caption{Ablation study of the search space and training strategy. 
}
\label{tab:compareablation}
\end{table}

\paragraph{Image Translation.} In terms of network level architecture, higher resolution is preferred by Pix2Pix architecture. The location of skip-connections searched by NAS method is essential. Since skip-connections that combining deep and shallow features is beneficial to the task of image translation. Besides, Pix2Pix model prefers longer distance connection compared with CycleGAN model since far skip connection converges less correlated information which is beneficial for textural details and overall reconstruction. In addition, as the network goes deeper, it is not necessary to increase the width. Only a few layers in the specific location should place large channel numbers, which minimize model size without losing performance. CF-GAN selects more DwsBlocks for the early layers; while leaving the last layers with Conv 3$\times$3. 

\paragraph{Super-Resolution.} We demonstrate the importance of the asymmetric pyramid structure with a progressive upsampling strategy from the searched architecture. Network architecture should have a deeper structure when the feature scale is relatively low. Otherwise, it should have a shallower structure. On the one hand, this can save the amount of calculation, on the other hand, it is beneficial to increase the receptive field of the original image and improve the quality of the image. 
In terms of block level architecture, GRB is often used in lower resolution while SRB is preferred at higher resolution, suggesting that different spatial resolutions are suitable for different structures.

\subsection{Ablation Study} 
\label{sec:ablationstudy}

\paragraph{Effectiveness of Search Space and Training Strategy.} 
\begin{figure}[t]
\centering
\begin{subfigure}[b]{0.236\textwidth}
	\includegraphics[width=\textwidth]{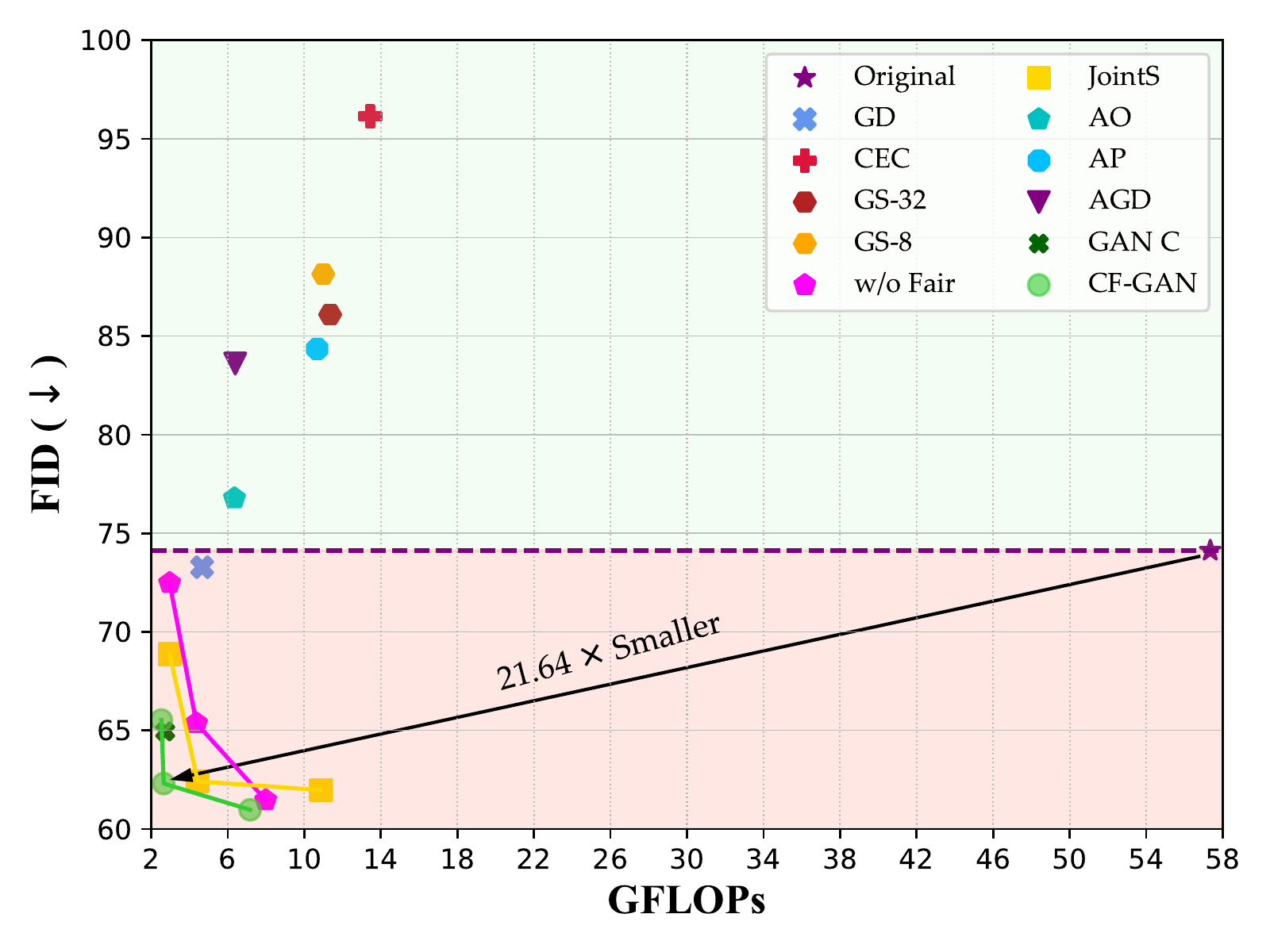}
	\caption{Horse $\to$ Zebra.}
	\label{fig:hor2zeb}
\end{subfigure}
\begin{subfigure}[b]{0.236\textwidth}
    \includegraphics[width=\textwidth]{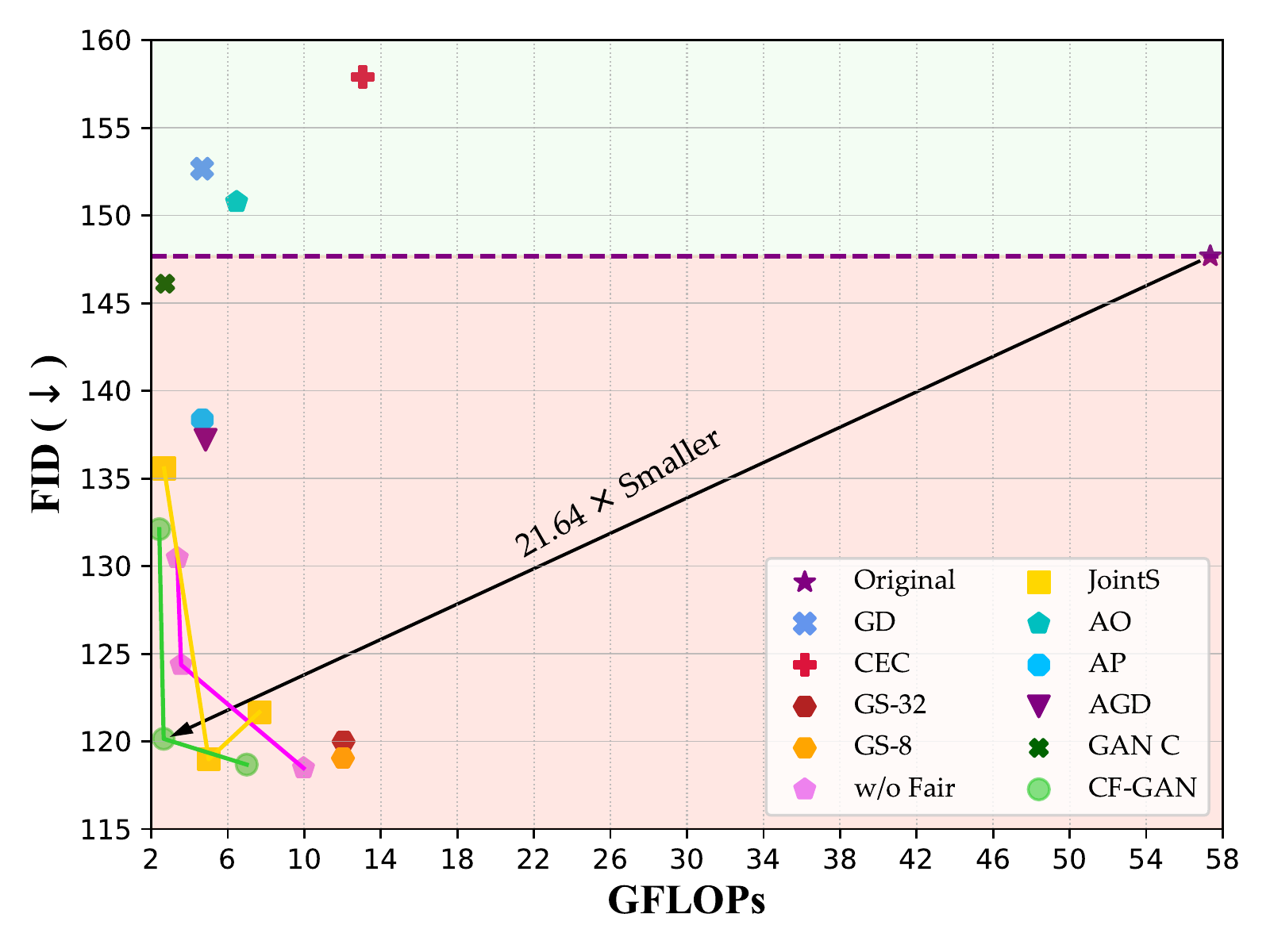}
	\caption{Zebra $\to$ Horse.}
	\label{fig:zeb2hor}
\end{subfigure}
\caption{Left: Numerical results of ablation studies on horse2zebra dataset. Right: Numerical results on zebra2horse dataset.}
\label{fig:compare}
\end{figure}

We conduct thorough ablation studies by comparing the following methods for CycleGAN. In order to show the superiority of our search space and search strategy, we conduct experiments on respectively replacing the search space of operator and path from automatic search to artificial design (\textit{AO, AP, Joint Search}). Specifically, \textbf{A} in Tab~\ref{tab:compareablation} stands for the present term is artificially design, while \textbf{N} stands for obtained by NAS. Besides, training strategies are also considered as controlled variables to explore their effects (\textit{w/o Fair Training}). Moreover, other SOTA optimization framework with pruning, distillation, quantization and their combinations are also considered (\textit{Distill, CEC, GS-32, GS-8, GAN Comp., AGD}). All involved methods are listed as follows:

\begin{itemize}
    \item Distillation (\ie, GAN Distillation \cite{chen2020distilling}).
    
    \item Co-Evolutionary pruning (\ie, CEC \cite{shu2019co}).
    
    \item GS-32 \cite{wang2020gan}: Jointly optimizing channel pruning, distillation and 32 bit quantization.
    
    \item GS-8 \cite{wang2020gan}: Jointly optimizing channel pruning, distillation and 8 bit quantization.
    
    \item w/o Fair: Directly pre-training in GAN's minimax manner without the \textit{Fair Super-GAN Training Strategy}.
    
    \item JointS: Jointly searching for optimal path, operators and channel, instead of coarse-to-fine searching.
    
    \item AO: Searching for optimal path and channel with artificial designed operators by one-shot NAS method.
    
    \item AP: Searching for optimal operators and channel with artificial designed path by one-shot NAS method.
    
    \item Distillation + searching for operators (\ie, AGD \cite{fu2020autogan}): Differential architecture search on operators and width.
    
    \item Distillation + searching for channel (\ie, GAN Compression\cite{gancomp_2020}): Only perform NAS on channel number search.
    
\end{itemize}

Numerical and visualization results are depicted in Tab~\ref{tab:compareablation} and Fig.~\ref{fig:compare} respectively. We find that any dimension change from automatic search to manual design would lead to a decrease in model performance and an increase in parameter quantity (\textit{AO, AP}). Meanwhile, the proposed fair training strategy also lead to better results (\textit{w/o Fair Training}). In addition, our CF-GAN achieves superior trade-off between image quality and model complexity with traditional distillation, pruning and quantization techniques and their combination (\textit{Distill, CEC, GS-32, GS-8}). Performing distillation with NAS (\textit{GAN Comp., AGD}) is another effective way to obtain efficient GAN models and can indeed improve image generation quality compared with single distillation. However, our methods outperform the method in the generation quality at similar compression ratio. 

\paragraph{Effectiveness of Coarse-to-fine Search.}
We compare the search time and metrics of joint search method and CF-GAN, as in Tab~\ref{tab:searchcost}. CF-GAN could save 7.2 $\sim$ 10$\times$ search time with pretty close model performance compared to the joint search method. Pix2pix, CycleGAN and ESRGAN models are respectively measured on horse2zebra, edges2shoes and Urban100 dataset.
\begin{table}[h]
\renewcommand*{\arraystretch}{1.15}
\setlength{\tabcolsep}{4.5pt}
\small\centering
\resizebox{0.47\textwidth}{!}{
\begin{tabular}{clcccc}
\toprule
\multirow{2}{*}[\multirowcenter]{Model} & \multirow{2}{*}[\multirowcenter]{Search method} & \multirow{2}{*}[\multirowcenter]{Search Cost}
& \multicolumn{1}{c}{\multirow{2}{*}[\multirowcenter]{GFLOPs}} &  
\multicolumn{2}{c}{Metric} \\ 
\cmidrule(lr){5-6}
& & & &\multicolumn{1}{c}{FID $\downarrow$} & \multicolumn{1}{c}{PSNR $\uparrow$} \\
\midrule
\multirow{2}{*}{Pix2Pix} & Joint Search & 62h & 4.96 & 24.10 & \multicolumn{1}{c}{--} \\
& CF-GAN & 8h & 4.77 & 24.13 & \multicolumn{1}{c}{--} \\
\cmidrule{1-6}
\multirow{2}{*}{CycleGAN} & Joint Search & 59h & 2.78 & 62.17 & \multicolumn{1}{c}{--} \\
& CF-GAN & 6h & 2.65 & 62.31 & \multicolumn{1}{c}{--} \\
\cmidrule{1-6}
\multirow{2}{*}{ESRGAN} & Joint Search & 72h  & 105.6 & \multicolumn{1}{c}{--} & 24.96  \\
& CF-GAN & 10h & 101.3 & \multicolumn{1}{c}{--} & 24.94  \\
\bottomrule
\end{tabular}
}
\vspace{5pt}
\caption{Comparison of search cost and compressed generator performance of Joint Search and Coarse-to-fine Search. 
}
\label{tab:searchcost}
\vspace{-5pt}
\end{table}

\section{Conclusion}
\label{sec:conclusion}
We propose \text{CF-GAN} for searching efficient network structures in GAN based models. \text{CF-GAN} automatically searches multiple dimensions for suitable compact architecture for versatile tasks within one unified coarse-to-fine optimization framework. An fair pre-training strategy for super-generator is thoroughly analysed and demonstrated to satisfy the equal treatment of all subnets. The searched models by CF-GAN outperforms existing GAN compression options with better generative quality, which provide inspiration for designing related GANs.

\clearpage
\appendix
The appendix is organized as follows: Appendix~\ref{sup:A_method} provides supplementary proofs of relevant theories and detailed version of the algorithm in the main paper. Appendix~\ref{sup:B_results} shows additional results of the searched GAN architecture and generated images.  

\section{Additional Method Details}
\label{sup:A_method}

\subsection{Proof of Limitations of Uniform Sampling}
In this section we prove that traditional uniform sampling method is not enough to meet the \textit{Fair Super-GAN Training} definition in Section 3.2 of the main paper, as an evidence of the necessity of our proposed \textit{Fair Training Strategy}. 

We define the uniform sampling as follows: 
For each step, we uniformly sample one path from $G_{\text{super}}$ without replacement dubbed $G_{p}$ and a corresponding discriminator path $D_{p}$ from $D_{\text{super}}$ to optimize. For a fair comparison, this step is controlled to remain the same as our proposed \textit{Fair Training Strategy}. After that, for $G_p$'s training process, we uniform sample a sub-generator from $M$ choices without replacement form $G_p$ to optimize. There exists a difference from our proposed \textit{Fair Training Strategy}, which breaks the \textit{Fair Super-GAN Training} definition. 
\begin{proof}
Here we assuming that uniform sampling satisfies \textit{Fair Super-GAN Training}. Denote $U_{p,l_{m}}^{t}$ as the times of updates over the first $t$ trials. $m$,$l$,$p$ are the index of operation, layer, and path respectively. Then accordingly for $\forall p\in [1,N_p], l\in [1,L]$, $t\in [1,T]$, we have:
\begin{equation}
p(U_{p,l_{1}}^{t}=U_{p,l_{2}}^{t}=\hdots=U_{p,l_{M}}^{t})=1
\label{proof:1}
\end{equation}

Here we consider a special case for simplicity: The $p$-th path is sampled for the first $t$ updates. Therefore, in the $t$ updates, a total update number of all blocks of the $p$-th path is $t$, which can be expressed as:
\begin{equation}
\sum_{i=1}^{M}{U_{p,l_i}^t}=t
\label{proof:2}
\end{equation}

Let $g(p,M,t)=p(U_{p,l_{1}}^{t}=U_{p,l_{2}}^{t}=\hdots=U_{p,l_{M}}^{t})$, considering Eq.~\ref{proof:2} we have:
\begin{equation}
g(p,M,t)=C_t^{\frac{t}{M}}C_{\frac{t(M-1)}{M}}^{\frac{t}{M}}\cdots C_{\frac{t}{M}}^{\frac{t}{M}}\frac{1}{M^t}=\frac{t!}{(\frac{t}{M}!)^{M}}\frac{1}{M^t}
\label{proof:3}
\end{equation}

Since $g(t)$ strictly decreases monotonically with $t$ and $g(t)>0$, its limitation exists. Then we calculate its limitation when $t\rightarrow + \infty$ based on Stirling approximation~\cite{tweddle2012james}:
\begin{equation}
\small
\begin{split}
\lim\limits_{t \to +\infty} g(p,M,t) &= \lim\limits_{t \to +\infty}\frac{t!}{(\frac{t}{M}!)^{M}\times M^t} \\
&= \lim\limits_{t \to +\infty} \frac{\sqrt{2\pi t}(\frac{t}{e})^t}{\sqrt{2\pi \frac{t}{M}}^{M} (\frac{t}{e})^t} \\
&= 0
\end{split}
\end{equation}
which means that $p(U_{p,l_{1}}^{t}=U_{p,l_{2}}^{t}=\hdots=U_{p,l_{M}}^{t})=0$ and leads to a contradiction.

In addition, the above proof can be easily extended to general cases in which each time a different path is sampled. 

\end{proof}

\subsection{Proof of Total Operator Specialization Number}
Given the pretrained supernet generator, we find the optimal path by the first step of our coarse-to-fine search strategy presented in Section 3.3 of the main paper. Suppose the searched path contains $L$ sequential layers, each with $M$ candidate choice blocks. Here we prove that the total operator specialization number is $N_o = (M!)^{L-1}$. 

\begin{proof}
When we uniform sample $M$ models without replacement, the number of all possible choices ($N_o$) can be formulated as:
\begin{equation}
    \label{eqn:choices}
     N_o = (L_1 * L_2 * \cdots * L_{M-1} * L_M)/A_M^M
\end{equation}
where $L_i$ denotes the number of possible samples of the $i$-th model. $A_M^M = M!$ represents the full  permutation of the $M$ samples. 

Since the sampling process is uniform sampling without replacement, the number of possibilities at the $i$-th  time can be written as:
\begin{equation}
    \label{eqn:sample}
     L_i = (M+1-i)^L
\end{equation}
Accordingly, Eq.~\ref{eqn:choices} can be expressed as:
\begin{equation}
    \label{eqn:choices}
     N_o = \prod_{i=1}^{M}\frac{(M+1-i)^L}{M!}=\frac{(M!)^L}{M!}=(M!)^{L-1}
\end{equation}

\end{proof}
Thus, in the second step of the progressive searching strategy, our goal is to find the optimal operator specialization that is from totally $N_o = (M!)^{L-1}$ choices.

\subsection{Algorithm of Equal Super-GAN Training}
We present the detailed version of algorithm of \textit{Fair Super-GAN Pre-training} as Algorithm~\ref{alg:ab_pretrain}. 
\begin{algorithm}[t]
\caption{Fair pretrain the supernet generator}
\label{alg:ab_pretrain}
\begin{algorithmic}[1]
\REQUIRE{training set $D_{train}$, super-generator $G_{\text{super}}$, super-discriminator $D_{\text{super}}$, pretrained generator $G_0$, total number of different paths ($N_{p}$), total number of choices for each layer ($M$), epochs to pretrain ($T_1$), learning rate $\left\{\alpha^{(t)}\right\}$, $\left\{\eta^{(t)}\right\}$.}\\
\STATE Initialize the super-GAN's weight $\{ w_G,w_D \}$ and architecture parameters ${\gamma}$ with uniform distribution. 
\FOR{$t$ $\gets$ 1\ to\ $T_1$ } 
\STATE Get data $(X,Y)$ from the training set $D_{tr}$. 
\FOR{$p$ $\gets$ 1\ to\ $N_p$ } 
\STATE Uniform sample $G_p$ from $G_{\text{super}}$ and corresponding $D_p$ from $D_{\text{super}}$ w/o replacement.
\STATE Clear gradients recorder for all weights.
\FOR{$m$ $\gets$ 1\ to\ $M$}
\STATE Uniform sample $G_o$ from $G_p$ w/o replacement for each block to build a sub-generator $G_o$. 
\STATE Calculate gradients of generator based on:\\
\quad $g_{G_o(w_G)}^{(t)} = \nabla_{w_{G}} L_{\text{train}}(G_o,G_0,D_p,X,Y)$
\STATE Accumulate gradients $g_{G_o(w_G)}^{(t)}$ to $g_{G_p(w_G)}^{(t)}$.
\ENDFOR
\STATE Update weights of $G_{p}$ by accumulated gradients.\\
\quad $w_{G}^{(t+1)} \gets w_{G}^{(t)} - \alpha^{(t)} g_{G_p(w_{G})}^{(t)}$
\STATE Calculate gradients of scale factors $\gamma$ based on:\\
\quad $g_{G_p(\gamma)}^{(t)} = \nabla_\gamma L_{\text{train}}(G_p,G_0,D_p,X,Y)$
\STATE Update scale factors of $G_p$ by accumulated gradients.\\
\quad $\gamma^{(t+1)} \gets \textrm{prox}_{\lambda_{sp}\eta^{(t)}} (\gamma^{(t)} - \eta^{(t)} g_{G_p(\gamma)}^{(t)})$ 
\STATE Calculate gradients of $D_{p}$ based on:\\
\quad $g_{D_p(w_D)}^{(t)} = \nabla_{w_{D}} L_{\text{train}}(G_p,G_0,D_p,X,Y)$
\STATE Updated weights of $D_{p}$:\\
\quad $w_{D}^{(t+1)}$ $\gets$ $w_{D}^{(t)} + \alpha^{(t)} g_{D_p(w_D)}^{(t)}$
\ENDFOR
\ENDFOR
\ENSURE{Pre-trained super-generator $G_{\text{pre}}$.}
\end{algorithmic}
\end{algorithm}

\subsection{Algorithm of Coarse-to-fine Searching}
We present the detailed version of algorithm for the whole CF-GAN method as Algorithm~\ref{alg:ab_CFGAN}. 
\begin{figure}[t]
\begin{minipage}{0.48\textwidth}
\vspace{-1.2em}
\begin{algorithm}[H]
\caption{The proposed CF-GAN method}
\label{alg:ab_CFGAN}
\begin{algorithmic}[1]
\REQUIRE{ training set $D_{train}$, validation set $D_{val}$, supernet $G_{\text{super}}$ and discriminator $D_{\text{super}}$, pretrained generator $G_0$, total number of different paths ($N_{p}$), total number of choices with different operators ($N_{o}$), fine-tune epochs ($T_2$), learning rate $\left\{\alpha^{(t)}\right\}$, $k$.}

\textit{\# First Step: Fairly Pretrain the Supernet Generator}

\STATE Obtain the pre-trained $G_{\text{super}}$ by Algorithm~\ref{alg:ab_pretrain} as $G_{\text{pre}}$.

\textit{\# Second Step: Coarse-to-fine Search}
\FOR{$n$ $\gets$ 1 \ to\ $N_p$ }
\STATE Sampling subnet $G_n$ from $G_{\text{pre}}$ without replacement that contains single-path but mixed operators. 
\STATE Inherit weights of $G_n$ directly from the pretrained supernet $G_{\text{pre}}$ without training from scratch. 
\STATE $f_n$ = Evaluate($G_n$, $D_{val}$)
\ENDFOR

\STATE Select the optimal architecture: $G_{\text{path}}={\underset {G_n}{{arg\,max} }}\,(f_n)$ and the corresponding $D_{\text{path}}$.
\FOR{$m$ $\gets$ 1 \ to\ $N_o$ }
\STATE Sampling subnet $G_m$ from $G_{\text{path}}$  without replacement that contains single-operator. 
\STATE Inherit weights of $G_m$ directly from $G_{\text{path}}$ without training from scratch. 
\STATE $f_m$ = Evaluate($G_m$, $D_{val}$)
\ENDFOR
\STATE Select the optimal architecture: $G_{\text{optr}}={\underset {G_m}{{arg\,max} }}\,(f_m)$
  
\STATE Obtain architecture $G_{\text{channel}}$ as $G^*$ from $G_{\text{optr}}$ by shrinking redundant channels with evolutionary algorithm. (Algorithm ~\ref{alg:genetic}).
  
\textit{\# Third Step: Fine-tune}
\FOR{$t$ $\gets$ 1\ to\ $T_2$ }
\STATE Get data $(X,Y)$ from the training set $D_{tr}$. 
\FOR{$k$ steps } 
\STATE Calculate gradients of generator based on:\\
\quad $g_{G^*(w_G)}^{(t)} = \nabla_{w_{G}} L_{\text{train}}(G^*,G_0,D_{\text{path}},X,Y)$
\STATE Update weights of $G^*$:\\
\quad $w_{G}^{(t+1)} \gets w_{G}^{(t)} - \alpha^{(t)} g_{G^*(w_{G})}^{(t)}$

\ENDFOR
\STATE Calculate gradients of discriminator based on:\\
\quad $g_{D_{\text{path}}(w_D)}^{(t)} = \nabla_{w_{D}} L_{\text{train}}(G^*,G_0,D_{\text{path}},X,Y)$
\STATE Updated weights of $D$:\\
\quad $w_{D}^{(t+1)}$ $\gets$ $w_{D}^{(t)} + \alpha^{(t)} g_{D_{\text{path}}(w_D)}^{(t)}$
\ENDFOR

\ENSURE{The compressed efficient generator $G^*$.}

\end{algorithmic}
\end{algorithm} 
\end{minipage}
\end{figure}

\subsection{Detailed Evolutionary Algorithm}
\begin{algorithm}[h]
	\caption{Evolutionary search for shrinking the channel numbers.}
	\label{alg:genetic}
	\begin{algorithmic}[1]
		 \STATE \textbf{Input:} supernet $G_3$, validation set $D_{val}$, population size $N$, elitism number $K$, max iteration $T$,mutation probability $r$ computational constraints $C$.
  \STATE \textbf{Output:} Searched architecture $G^*$ under computational constraints.
		\STATE Initialize populations $P_0$ so that $|P_0| = N_{pop}$ and $|P_0|$ satisfies constraints $C$; elitism $E_0= \emptyset $
		\FOR{$t = 1$ to $T$}
		\STATE ${\text{Fitness}_{t-1}}={\text{Inference}}(G_3, P_{t-1}, D_{val})$
		\STATE Elitism $E_t=\text{Select}(\left\{ E_{t-1}, P_{t-1}\right\}, \text{Fitness}_{t-1},K)$

		\FOR{$i = 1$ to $\frac{N}{2}$}
		\STATE $\text{children}_i=\text{Crossover}(E_{t-1},C)$
		\STATE $\text{children}_{i+\frac{N}{2}}=\text{Directional\_Mutate}(E_{t},r,C)$
		
		\ENDFOR
		\STATE $P_t=\text{children}$
		\ENDFOR
		\STATE Update finesses of individuals in $P_{T}$;
		\STATE Establish searched generator network $G^*$ by exploiting to the best individual in $P_T$;
		\RETURN Searched generator $G^*$.
	\end{algorithmic}
\end{algorithm}
The searching procedure is confronted with the challenge of massive search cost, especially in image translation task or super resolution task since feature maps are computed in the same scale or higher resolution scale. Therefore, we resort to advanced evolutionary algorithm  with directional mutation to alleviate the issue. Inspired by the block credit guided evolutionary process~\cite{song2019efficient}, we propose an efficient approach to accelerate convergence  by taking into account significance of channel configuration in the whole architecture to direct the mutation. 

The evolutionary algorithm was used in the "Auto Channel Shrinking" step. Given searched generator $G_{\text{optr}}$ obtained by the previous step, we maintain a population size of $N_{pop}$ and select the top $K$ elites as elitism (or parents) according to the fitness of population and evolve it cyclically.
The next generation is produced half by crossover and half by mutation. Among them, the parents of the crossover process are from the last generation's chosen elite. For mutation, common methods randomly select a block and randomly change the channel number to replace an old architecture with certain probability to produce a new candidate. Despite the well-performance in traditional classification task, the undirected mutation process will result in time-consuming search and meaningless energy consumption in our pixel-level task. 

We modify the classical method by proposing \emph{Directional Mutation} method, where each channel configuration during mutation is proportional to the value of replacing with this width. We define \emph{Replacement Gain} (RG), which is calculated as follows:

\begin{equation}
    \label{eqn:credit}
    \text{RG}(c_j,l)=y_{c_j,l}^{\text{rep}}-y_{c_j,l}^{\text{bef}},
\end{equation}
where $y_{c_j,l}^{\text{rep}}$ represents the fitness after replacing the channel number of the feature map in layer $l$ with $c_j$ while $y_{c_j,l}^{\text{bef}}$ represents the fitness before replacement. Hence $\text{RG}$ denotes the fitness gains if we mutate the channel configuration $c_i$ in a certain layer to $c_j$. To enhance the diversity of different blocks, we normalize RG as follows:

\begin{equation}
    \label{eqn:normalize}
    \text{RG}_\text{n}(c_j,l)=\text{RG}(c_j,l)-(\min_l\left\{ \text{RG}(c_j,l) \right\}+\epsilon),
\end{equation}
where $\epsilon$ is a tiny constant and $\text{RG}_\text{n}$ represents normalized replacement gain. When the mutation process occurs, the probability of replacing the channel number of the $l$ layer feature with $c_j$ is:

\begin{equation}
    \label{eqn:mutates}
    P_{\text{select}}(c_j|l=l)=\frac{\text{RG}_\text{n}(j,l)}{\sum_{j=1}^{N_c}{\text{RG}_\text{n}(j,l)}},
\end{equation}
where $N_c$ denotes the number of width choices. 

\begin{figure}[t]
\vspace{5pt}
\centering
\begin{subfigure}[b]{0.235\textwidth}
	\includegraphics[width=\textwidth]{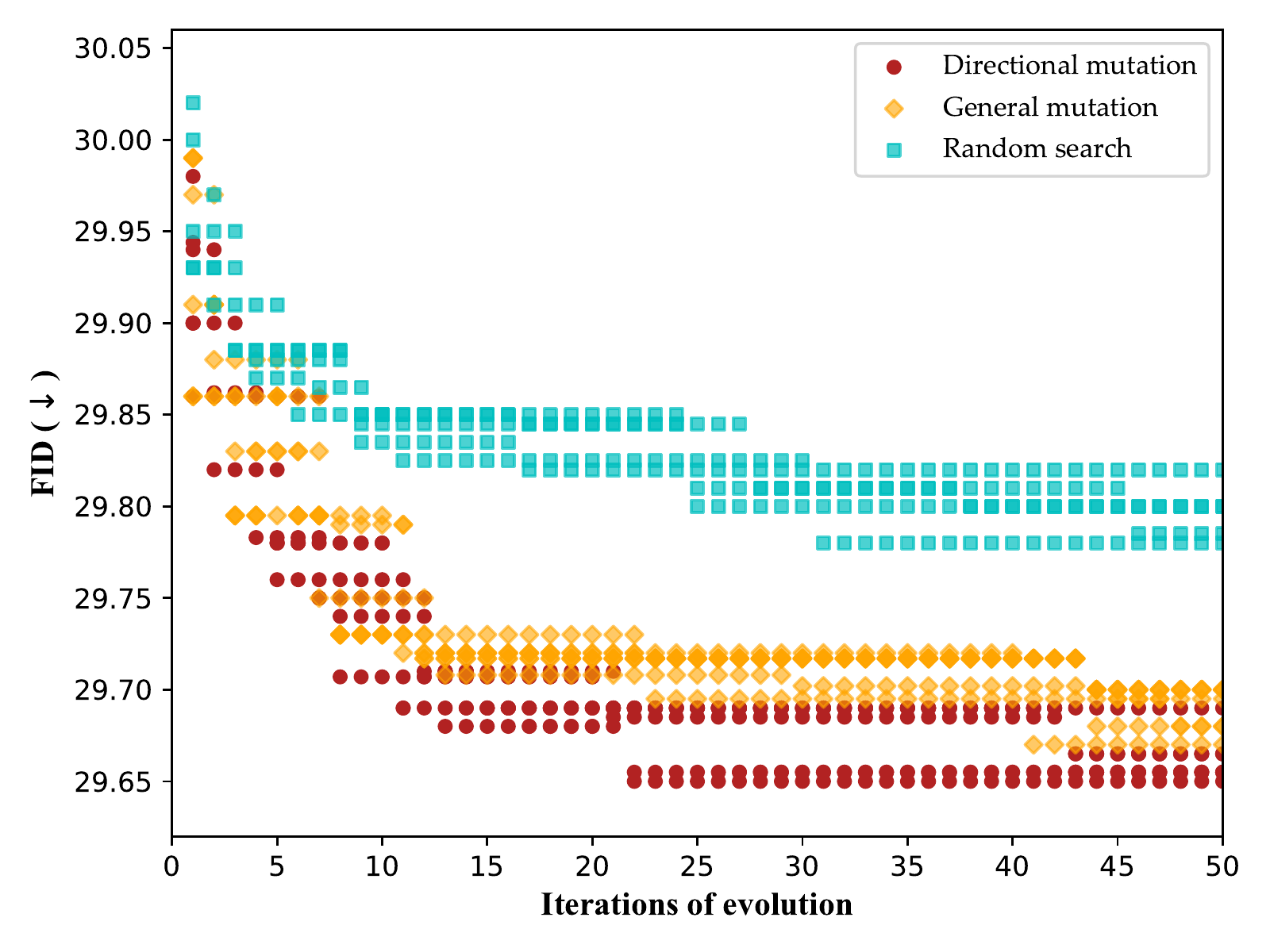}
	\caption{Image Translation.}
	\label{fig:imaget}
\end{subfigure}
\begin{subfigure}[b]{0.235\textwidth}
    \includegraphics[width=\textwidth]{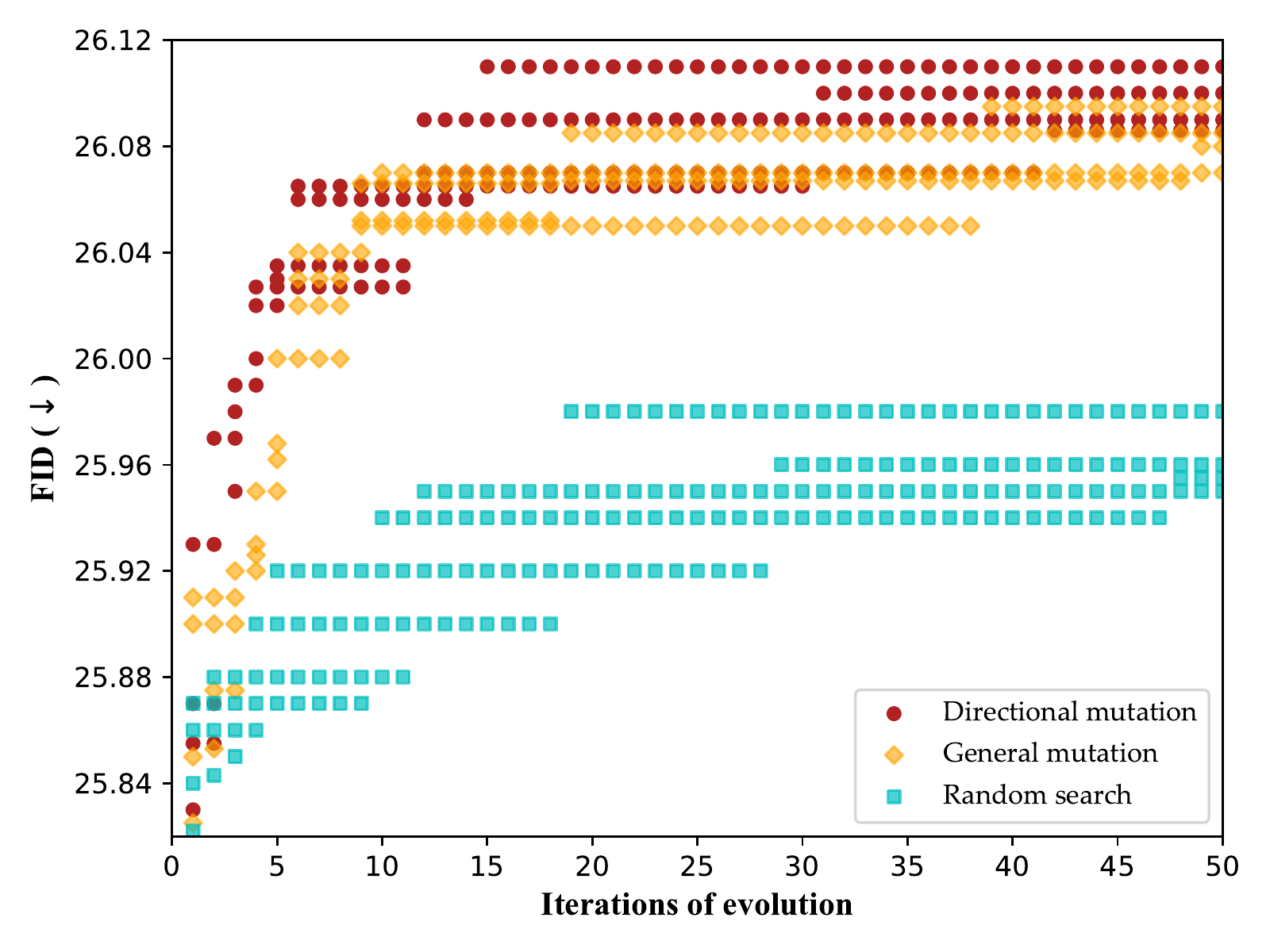}
	\caption{Super Resolution.}
	\label{fig:superr}
\end{subfigure}
\caption{Numerical comparison among our directional mutation, traditional mutation and random search.}
\label{fig:evo}
\vspace{-15pt}
\end{figure}

Directional Mutation strategy favors the channel configuration with higher performance gain in search space, which accelerates the convergence speed and actively aims to find better architecture. We repeat crossover and mutation process to generate enough new candidates. Our goal is to find the best channel configuration with constraint on GFLOPs and parameters, which is depicted as follows:
\begin{equation}
\label{single_obj} 
\begin{split}
\begin{aligned}
& \max_{conf \in S} \ \text{Fitness}(conf) \\
& \text{s.t}. \ \ \text{Param.}(conf) < P_{t}, \\
&\quad \ \ \ \  \text{FLOPs}(conf) < F_{t}
\end{aligned}
\end{split}
\end{equation}
where $conf$ denotes the channel configuration and $P_t$ and $F_t$ respectively represents the model size constraints and computation constraints. We present the detailed version of algorithm for our evolutionary algorithm with directional mutation as Algorithm~\ref{alg:genetic}. The application of  different mutation  strategies  for  searching  the  sub-generator of image translation and super resolution is conducted on the benchmarks. The results (See Figure~\ref{fig:evo}) show that,  compared with naive approaches, the proposed method can not only accelerates evolutionary speed but also  find a more satisfactory network architecture. It is mainly because the direction variation takes into  account the gain of various configurations, thus inducing the algorithm to obtain better configurations. 

\subsection{Detailed Method of Updating Scale Factor}
The scale factor in Algorithm ~\ref{alg:ab_pretrain} is used to decide which unimportant channels to prune after obtaining the optimal channel number configuration. We set a threshold according to the optimal channel number, and channels with scale factor value lower than this threshold will be pruned. We update the non-differentiable parameters $\gamma$ by performing proximal gradient descent to the results obtained by gradient descent. which can be presented as follows:
\begin{align} 
    g_{\gamma}^{(t)} &\gets \nabla_{\gamma} L_{\gamma}(\gamma)\bigg\rvert_{\gamma=\gamma^{(t)}} \\
    \gamma^{(t+1)} &\gets \textrm{prox}_{\lambda_{sp}\eta^{(t)}}(\gamma^{(t)}-\eta^{(t)}g_{\gamma}^{(t)})
\end{align}
where $\gamma^{(t)}$ and $\eta^{(t)}$ are the values of $\gamma$ and learning rate at step $t$, respectively. $L_{\gamma}$ denotes terms containing $\gamma$ in the objective function $\mathcal{L}_\text{train}$. The proximal function $\textrm{prox}_{\lambda\eta}(\cdot)$ for the $\ell_1$ constraint is:

\begin{equation}
    \label{eqn:prox}
    \textrm{prox}_{\lambda\eta}(\text{s}) = \begin{cases}
        \text{s}- \lambda\eta & \text{if} \ \text{s} > \lambda\eta, \\
        \mathbf{0} & \text{if}\ \lvert \text{s} \rvert \leq \lambda\eta,    \\
        \text{s}+ \lambda\eta & \text{if}\ \text{s} < -\lambda\eta. 
        \\
    \end{cases}
\end{equation}
where scalar $\text{s}$ can be regarded as each dimension of the input tensor.

\section{Additional Results}
\label{sup:B_results}

\subsection{Detailed Searched Architecture}
\label{app:architecture}

\begin{figure*}[t]
\centering
\begin{subfigure}[b]{0.88\textwidth}
	\includegraphics[width=\textwidth]{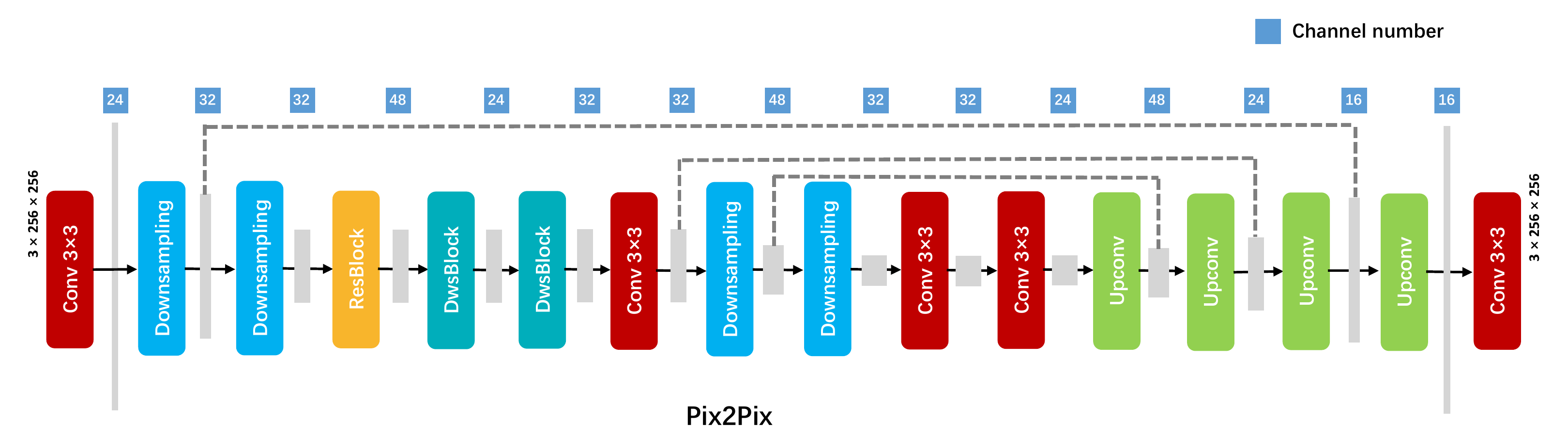}
	\caption{Searched Pix2Pix architecture}
	\label{fig:ab_searched_pix}
\end{subfigure}
\vspace{5pt}
\begin{subfigure}[b]{0.88\textwidth}
    \includegraphics[width=\textwidth]{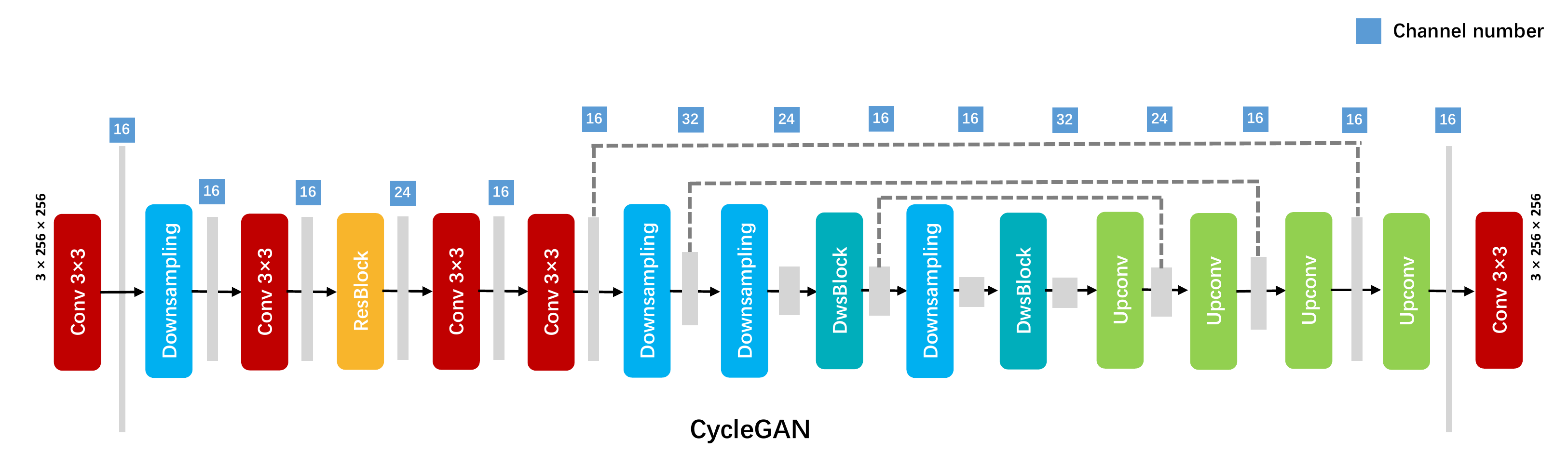}
	\caption{Searched CycleGAN architecture}
	\label{fig:ab_searched_cycle}
\end{subfigure}

\vspace{5pt}

\begin{subfigure}[b]{0.88\textwidth}
    \includegraphics[width=\textwidth]{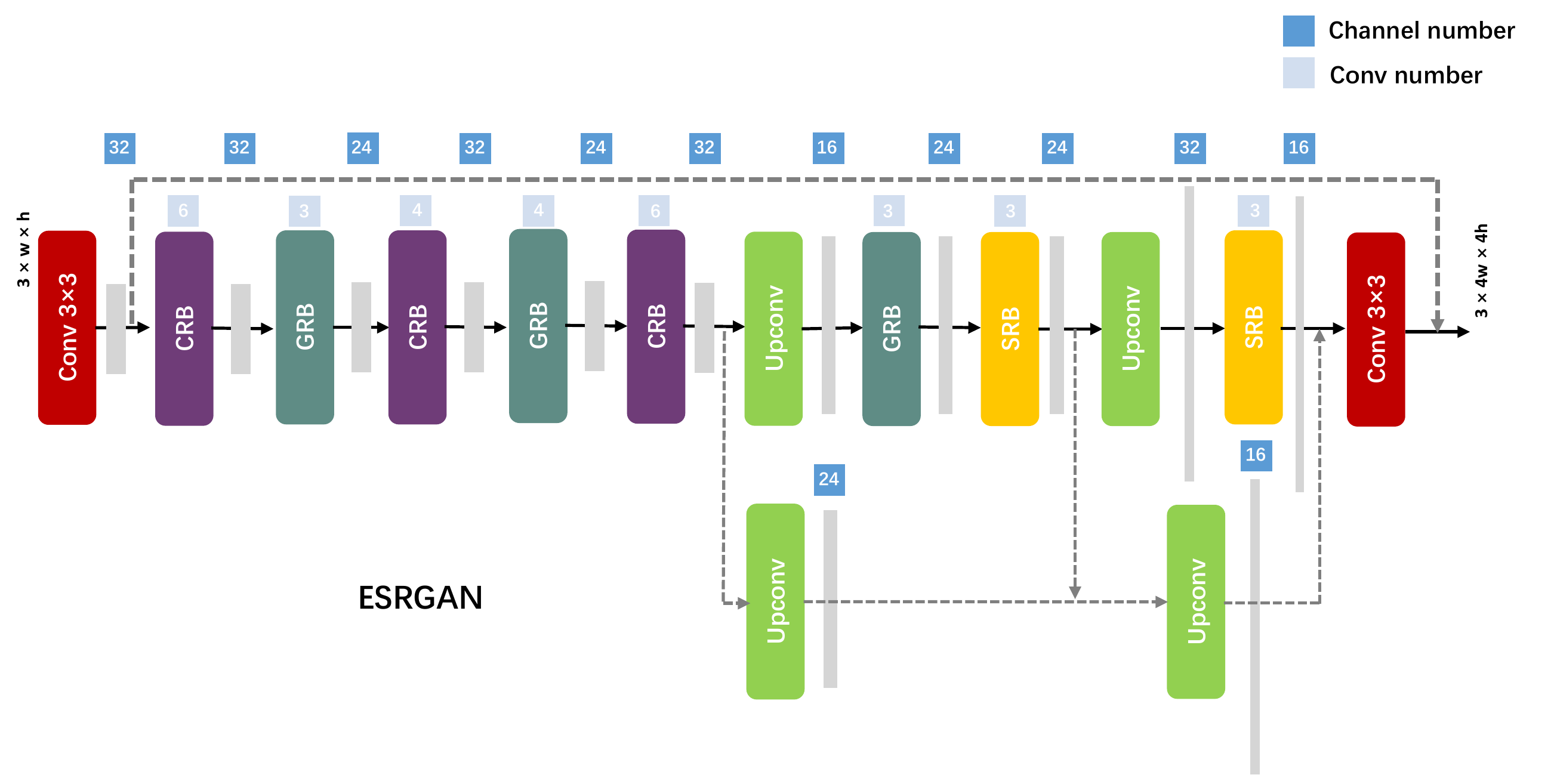}
	\caption{Searched ESRGAN architecture}
	\label{fig:ab_searched_esr}
\end{subfigure}
\caption{Illustration of the network architecture searched by CF-GAN respectively on different baselines: Pix2Pix (Top), CycleGAN (Middle) and ESRGAN (Bottom). We further leverage  sparsity regularization to retain important channels in each feature map. One can see that as the network goes deeper, it is not necessary to increase the width. Only a few layers in the specific location found by NAS should place heavy channel numbers, which is in line with intuition.} 
\label{fig:ab_searched_all}
\end{figure*}

Illustration of the network architecture searched by CF-GAN is shown in Figure~\ref{fig:ab_searched_all}.

\subsection{Additional Visual Results}
\begin{figure*}[t]
\centering
\includegraphics[width=\linewidth]{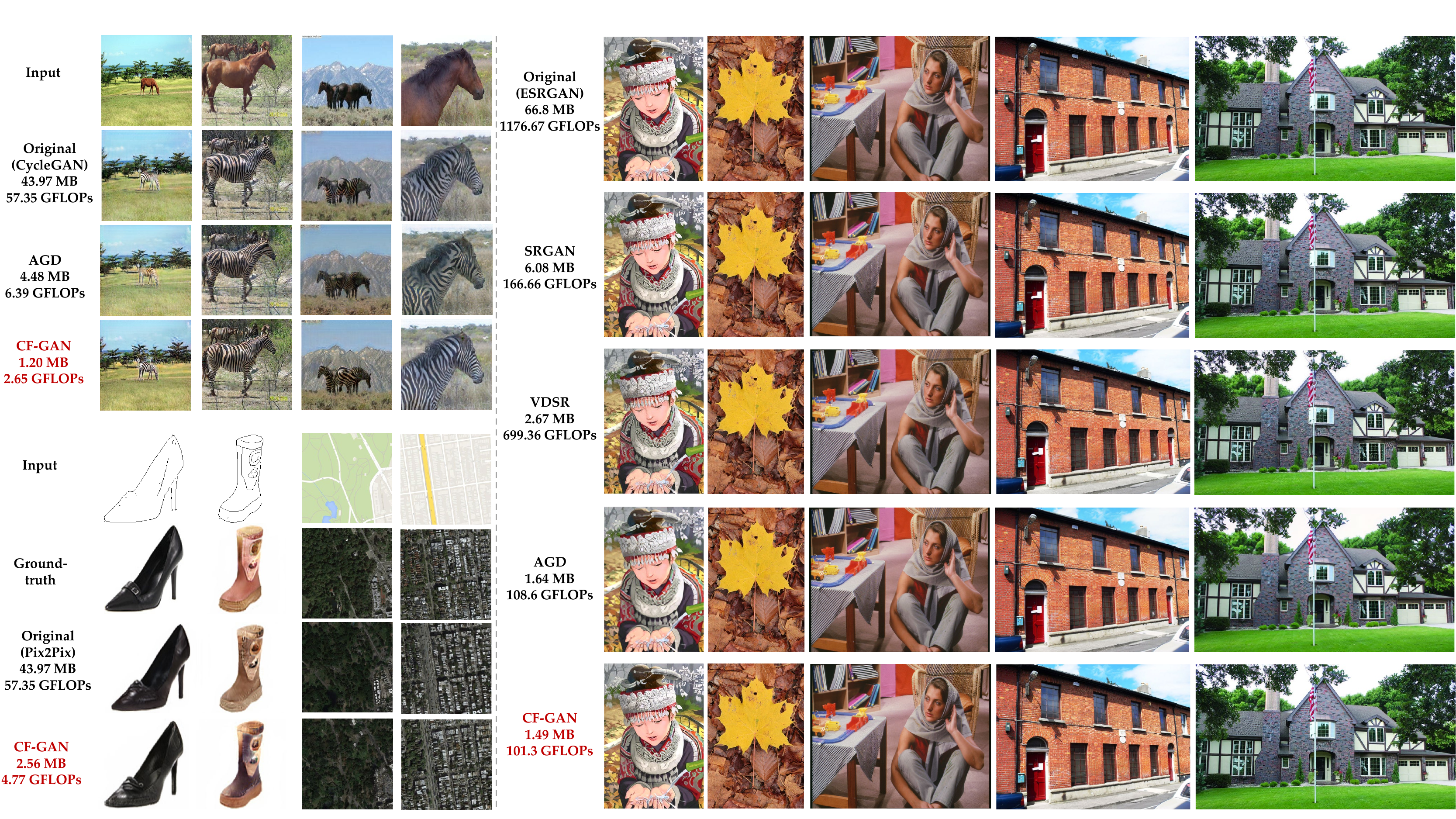}
\caption{Additional qualitative results of CF-GAN on different super resolution methods, CycleGAN and Pix2Pix compression with benchmarks (Part \uppercase\expandafter{\romannumeral1}).}
\label{fig:quali1}
\end{figure*}

\begin{figure*}[t]
\centering
\includegraphics[width=\linewidth]{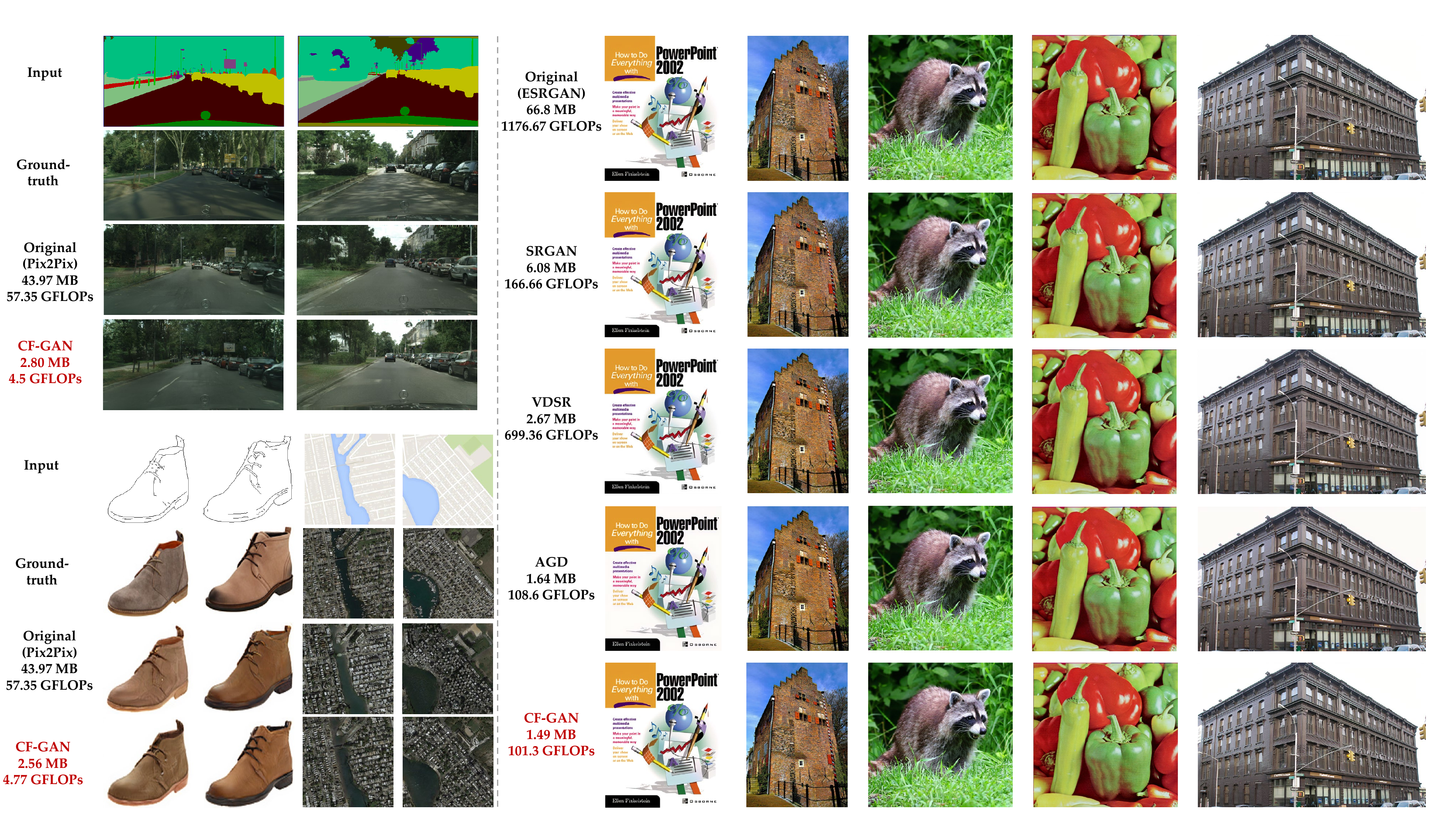}
\caption{Additional qualitative results of CF-GAN on different super resolution methods, CycleGAN and Pix2Pix compression with benchmarks (Part \uppercase\expandafter{\romannumeral2}).}
\label{fig:quali2}
\end{figure*}

\begin{figure*}[t]
\centering
\includegraphics[width=\linewidth]{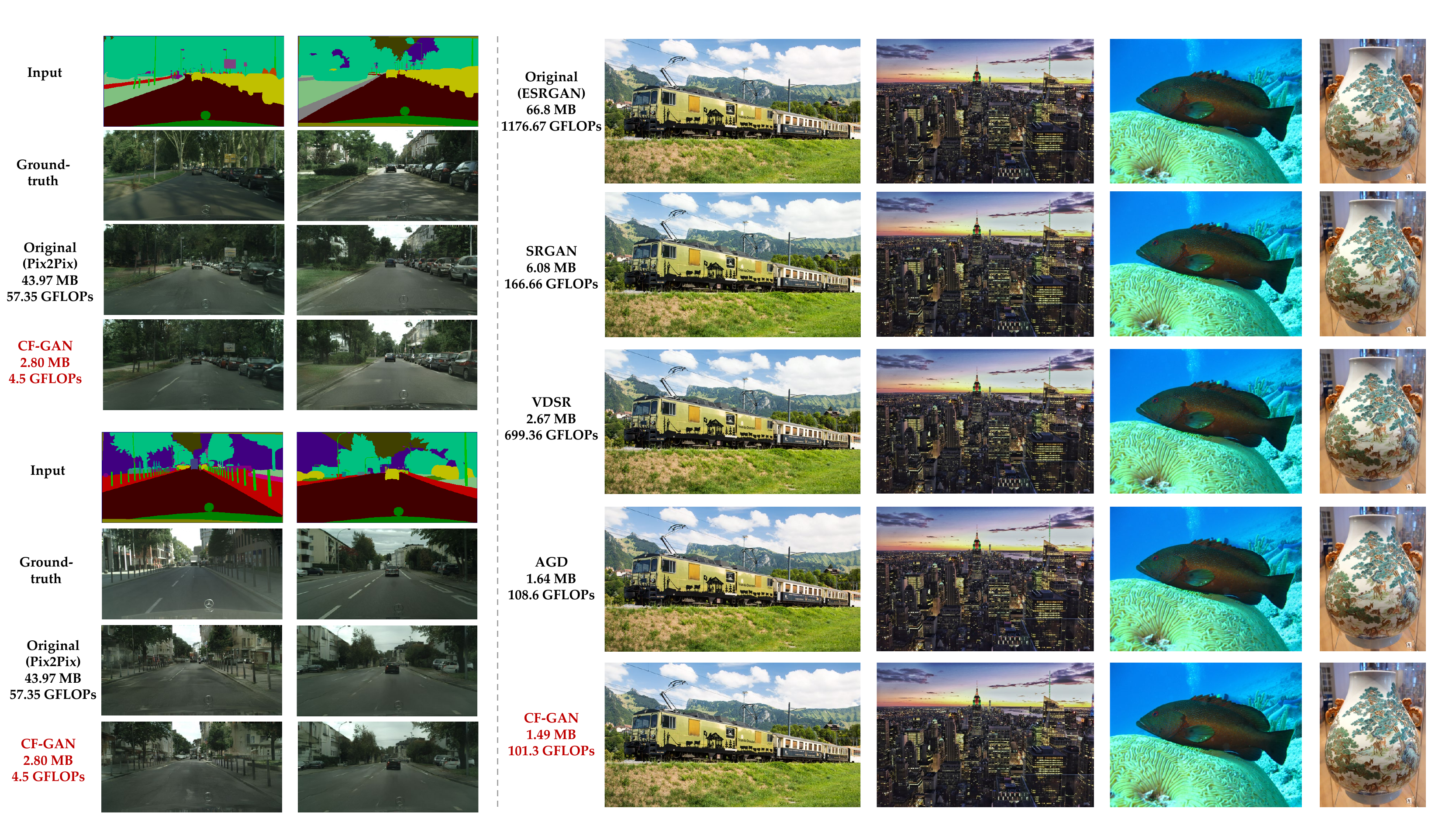}
\caption{Additional qualitative results of CF-GAN on different super resolution methods, CycleGAN and Pix2Pix compression with benchmarks (Part \uppercase\expandafter{\romannumeral3}).}
\label{fig:quali3}
\end{figure*}

\begin{figure*}[t]
\centering
\includegraphics[width=\linewidth]{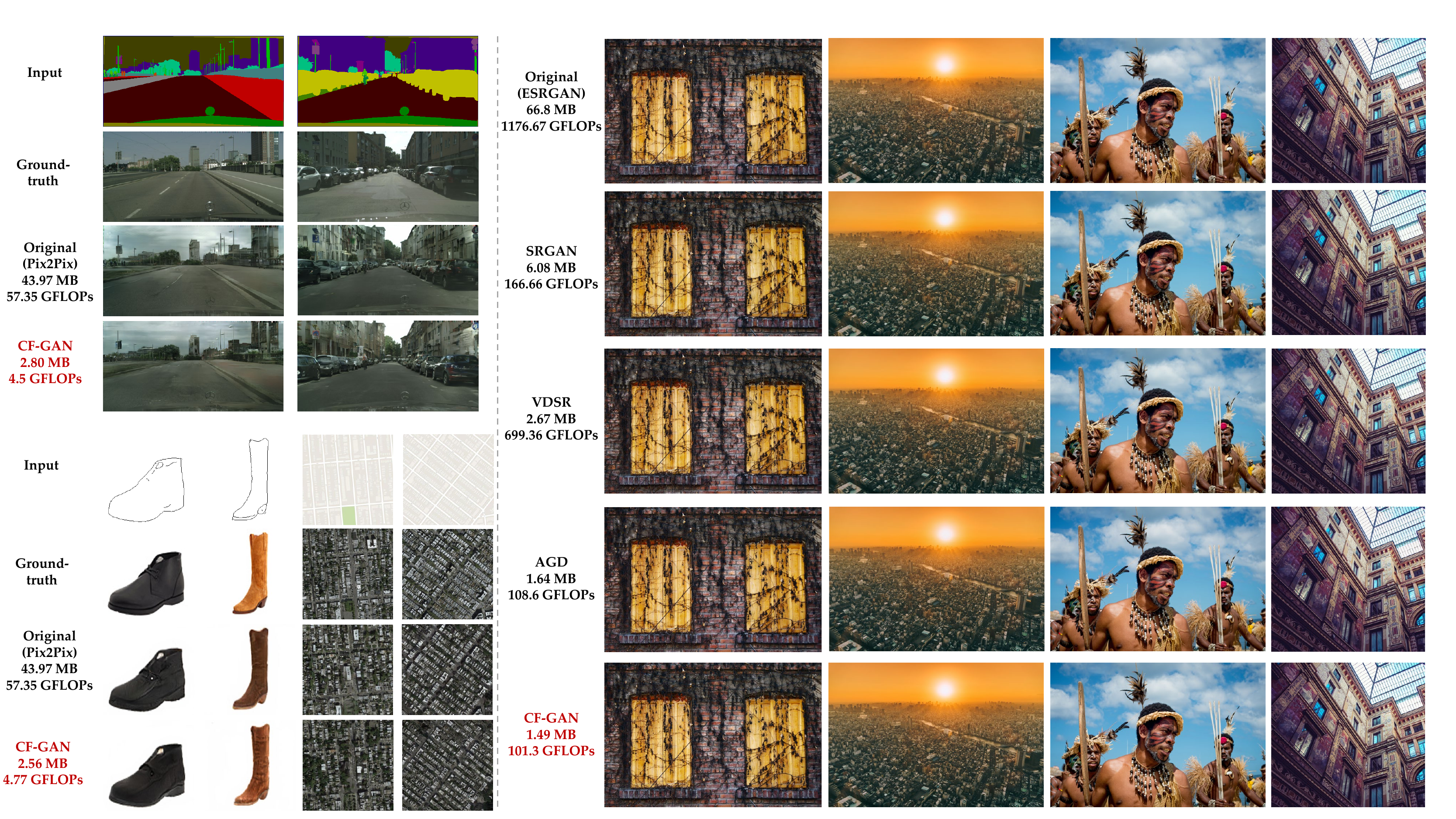}
\caption{Additional qualitative results of CF-GAN on different super resolution methods, CycleGAN and Pix2Pix compression with benchmarks (Part \uppercase\expandafter{\romannumeral4}).}
\label{fig:quali4}
\end{figure*}

In Fig.\ref{fig:quali1}, Fig.\ref{fig:quali2}, Fig.\ref{fig:quali3} and Fig.\ref{fig:quali4} we show additional visual results of our proposed CF-GAN method on image translation and super resolution tasks with the benchmarks.
\label{app:architecture}
\clearpage
{\small
\bibliographystyle{ieee_fullname}
\bibliography{main}
}

\end{document}